\documentclass[journal,twocolumn]{IEEEtran}
\usepackage{cite}
\usepackage{amsmath,amssymb,amsfonts}
\usepackage{algorithmic}

\usepackage{xcolor}
\usepackage{pgfplots}
\usepgfplotslibrary{fillbetween}
\usepackage{pgfplotstable}
\pgfplotsset{compat=1.17}
\usepackage{tikz}

\usepackage{graphicx}
\usepackage{textcomp}
\usepackage{longtable}
\usepackage{array}
\usepackage{tabularx}
\usepackage{multirow}
\usepackage{hyperref}
\usepackage[final]{changes}
\usepackage{rotating}
\usepackage{pifont}
\newcommand{\cmark}{\ding{51}}%
\newcommand{\xmark}{\ding{55}}%

\usepackage{textcomp}
\usepackage{stfloats}
\usepackage{url}
\usepackage{verbatim}

\def\BibTeX{{\rm B\kern-.05em{\sc i\kern-.025em b}\kern-.08em
    T\kern-.1667em\lower.7ex\hbox{E}\kern-.125emX}}
\graphicspath{{./figures/}}
\begin{document}

\title{TRACE: Transformer-based Risk Assessment for Clinical Evaluation}
\author{Dionysis Christopoulos,
Sotiris Spanos, Valsamis Ntouskos, and Konstantinos Karantzalos
\thanks{This work was supported by the iToBoS EU H2020 project under Grant 965221.}
\thanks{D. Chirstopoulos, S. Spanos and K. Karantzalos are with the Remote Sensing Lab, National Technical University of Athens, Greece. V. Ntouskos is with the Department of Engineering and Sciences, Universitas Mercatorum, Rome, Italy.}
}

\markboth
{Christopoulos \MakeLowercase{\textit{et al.}}: Transformer-based Risk Assessment for Clinical Evaluation}
{Christopoulos \MakeLowercase{\textit{et al.}}: Transformer-based Risk Assessment for Clinical Evaluation}

\maketitle

\begin{abstract}
\small
We present TRACE (Transformer-based Risk Assessment for Clinical Evaluation), a novel method for clinical risk assessment based on clinical data, leveraging the self-attention mechanism for enhanced feature interaction and result interpretation. Our approach is able to handle different data modalities, including continuous, categorical and multiple-choice (checkbox) attributes. The proposed architecture features a shared representation of the clinical data obtained by integrating specialized embeddings of each  data modality, enabling the detection of high-risk individuals using Transformer encoder layers. To assess the effectiveness of the proposed method, a strong baseline based on non-negative multi-layer perceptrons (MLPs) is introduced. The proposed method outperforms various baselines widely used in the domain of clinical risk assessment, while effectively handling missing values. In terms of explainability, our Transformer-based method offers easily interpretable results via attention weights, further enhancing the clinicians' decision-making process.
\end{abstract}

\begin{IEEEkeywords}
self-attention, transformer encoder, tabular data, clinical data, melanoma, heart attack, risk estimation, computer aided diagnosis
\end{IEEEkeywords}


\section{Introduction}
\label{sec:introduction}

Healthcare is an industry that can gain significantly by utilizing modern artificial intelligence (AI) and machine learning (ML) methods by assisting clinicians in diagnosis, risk assessment and pathology of diseases. By incorporating AI-driven risk assessment algorithms (\cite{yin2021role, SHRIVASTAVA20179, giordano2021accessing}), clinicians can offer risk stratification of the patients for screening recommendations, which can help in early detection of diseases, enabling better informed decision-making from the clinicians, on the one hand, and timely intervention for patients that are considered high-risk, on the other. 

Healthcare data, typically collected using questionnaire-based surveys, exhibit a large diversity both in the nature, the quantity and the completeness of the attributes that are recorded (or reported) for each patient. Importantly, clinical data are multi-modal, comprising a combination of numerical (e.g., age, height, weight etc.) and categorical features (i.e., eye color, hair color, etc.) or even ``checkboxes'', where multiple values within the same feature are valid simultaneously (e.g., ancestry, doctors visited, etc.). However, missing values and other issues affecting clinical, and tabular data in general, also pose a significant challenge to maximize data utilization. This is crucial in the case of healthcare data, as they are generally scarce and their collection is laborious and with high cost, while AI-based methods, and deep-learning in particular, typically assume large quantities of data for training representative models that generalize well. Finally, another characteristic of clinical data is the notable imbalance between case and control groups.

These aspects, necessitate the development of task-specific AI and ML methods for the clinical domain. To address the data scarcity, effective methodologies dealing with clinical data should either enhance the data artificially (e.g., through data augmentation/imputation, generative models, etc.) or introduce ways to effectively utilize samples suffering from missing values, inconsistencies, and other data problematics. In this work, we propose a transformer-based \cite{NIPS2017_Transformers} clinical risk assessment model covering the spectrum of clinical data feature modalities that explicitly handles missing values, leading to improved performance with minimal computational overhead. Another key contribution of our work is the explainability provided through the generation of attention maps, which facilitates the  interpretability of the produced results both for computer scientists and for clinicians.

In summary, the contributions of this work are summarized below:

\begin{itemize}
    \item We propose a novel framework to perform clinical risk assessment using different data modalities that explicitly handles instances with missing values.
    \item We introduce ``checkbox embeddings'' to handle features with multiple valid categories simultaneously, commonly extracted from questionnaire-based surveys.
    \item The proposed model offers improved explainability, assisting clinicians in interpreting the model results and taking informed decision.
    \item We introduce a low-parameter non-negative neural network inspired by \cite{Rieckmann2022}, optimized for efficient clinical risk assessment, while being easily interpretable.
    \item The proposed TRACE architecture automatically handles missing values in both continuous and categorical features, without requiring additional imputation steps.
    \item We perform extensive experiments on six clinical datasets for skin cancer, (Melanoma,  Basal Cell Carcinoma \& Squamous Cell Carcinoma) heart disease and diabetes risk assessment, showing that the proposed model, achieves competitive performance with respect to the state-of-the-art. 
\end{itemize}

The source code of the proposed methodology is made publicly available at \url{https://github.com/DionysisChristopoulos/TRACE}.

\section{Related Work}
\label{sec:related}
Risk assessment models have been developed in the healthcare domain, providing risk scores for health condition diagnoses, based on personal clinical data. Logistic regression is a widely used method for such tasks (i.e., primary melanoma classification)\cite{Cho2005, Vuong2020}. Another study\cite{Xie2019} employs various machine learning classifiers like Decision Trees, Polynomial/Rbf/Linear SVM, Naive Bayes, Random Forest and Logistic Regression, as well as a neural network model, aiming to predict type 2 diabetes risk and identifying associated risk factors. Regarding the predictive accuracy, the neural network model, as expected, achieved the best results, suggesting that deep learning models can potentially develop critical decision-making abilities in medical tasks, to enhance prevention on various healthcare chronic conditions. Similarly, \cite{Weng2017} employs a single hidden layer MLP network for cardiovascular disease risk prediction, although stating its weak interpretability.

Recent works such as \cite{Rieckmann2022, dauchelle2024constrained} utilize the properties of ``non-negative'' neural networks, demonstrating their effectiveness on exploring various combinations of potential causes on health outcomes, based on clinical data, or increase the interpretability of medical image analysis, respectively. The architecture proposed in \cite{Rieckmann2022} constraints all learnable weights to non-negative values in order to ensure that the existence of an exposure can only increase the risk of the final outcome. In contrast with the weights, biases are constrained to be negative, acting as a threshold that only allows large values to go through the activation function and subsequently affect (positively) the final outcome. If a person ends up having no risk contribution to any of the exposures, meaning that the outputs across every node were zeros, then it is assumed that this person has a risk equal to a predefined baseline risk. In our work, we design an improved non-negative neural network as a strong baseline for the task at hand.

Extending the task in other domains, many works shifted towards leveraging Transformer architectures \cite{NIPS2017_Transformers}, originally designed for Natural Language Processing tasks, to handle tabular data. This shift is motivated through the self-attention mechanism, which provides the Transformer model with the ability to capture complex feature interactions and dependencies. TabTransformer\cite{huang2020tabtransformer} developed a framework that is composed of a column embedding layer to transform categorical features into learnable vectors. FT-Transformer\cite{Gorishniy2021RevisitingDL} improves on the previous work, as it deploys a Feature Tokenizer, converting both categorical and numerical features into learnable embeddings. However, there are limitation on these methods such as the handling of missing values. In such cases, Tab-Transformer uses the average of the learned embeddings of all classes within the corresponding column with the missing entries.

DANet \cite{DANET} introduces an Abstract Layer (ABSTLAY) that learns to group correlated features using learnable sparse masks and abstracts higher-level features from these groups, stacking ABSTLAYs forms a deep network that recursively captures global semantics while incorporating shortcut paths and structure re-parameterization to reduce inference complexity.

Similarly, GATE \cite{GATE} employs a novel gating mechanism, motivated by GRU, to perform feature representation learning with built-in feature selection, and then combines an ensemble of differentiable non-linear decision trees, whose outputs are re-weighted via self-attention, effectively blending decision tree inductive biases with deep learning for enhanced efficiency and performance.

GANDALF \cite{GANDALF} adapts GRU-inspired gating into a non-temporal, stage-wise Gated Feature Learning Unit (GFLU) that uses learnable masks to select and refine features over multiple stages, aggregating the resulting representations via a simple MLP to achieve interpretability and competitive accuracy with reduced hyperparameter tuning.

Lastly, TabNet \cite{TabNet} uses sequential attention for instance-wise feature selection, dynamically choosing which features to process at each decision step, through an encoder–decoder architecture that leverages sparse masks and attentive transformers to efficiently learn from raw tabular data while providing both local and global interpretability, with the added benefit of unsupervised pre-training.

Additionally, baseline methods, in order to address class imbalance, utilize oversampling techniques such as SMOTE \cite{Chawla_2002} and CTGAN \cite{ctgan} to generate synthetic samples on the minority class. However, in our approach, we address class imbalance implicitly by employing the Focal loss, which reduces the need for synthetic data generation.

\begin{figure}[t!]
\begin{center}
\begin{tabular}{c} 
   \includegraphics[width=0.9\columnwidth]{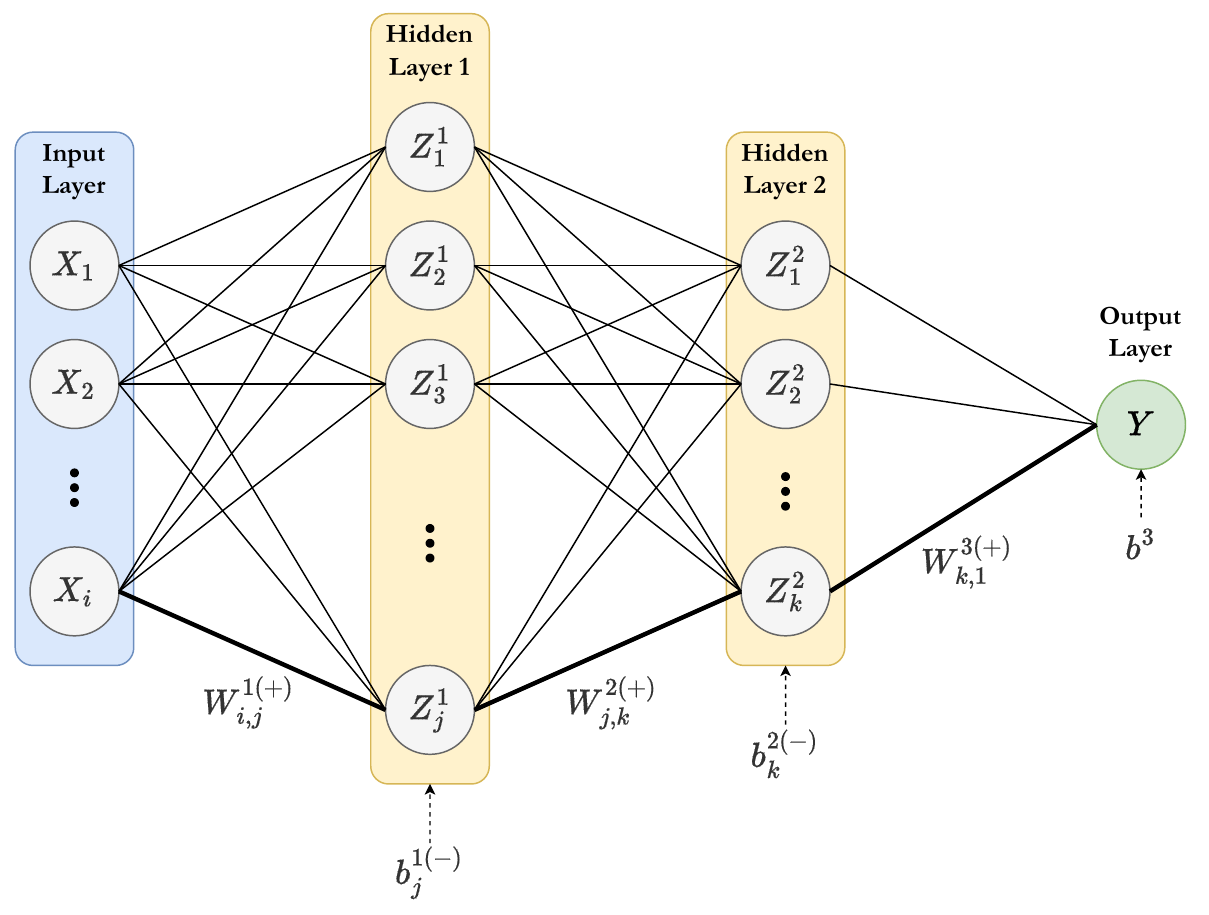}
\end{tabular}
\end{center}
\caption{Non-negative MLP architecture. The weight matrices $W^{1,2,3 (+)}$ are constrained to non-negative values, and the biases $b^{1,2 (-)}$ of the two hidden layers to negative values. Bias $b^3$ is left unconstrained.}
\label{fig:nnmlp}
\end{figure}

\begin{figure*}[t!]
\begin{center}
\begin{tabular}{c} 
   \includegraphics[width=0.8\textwidth]{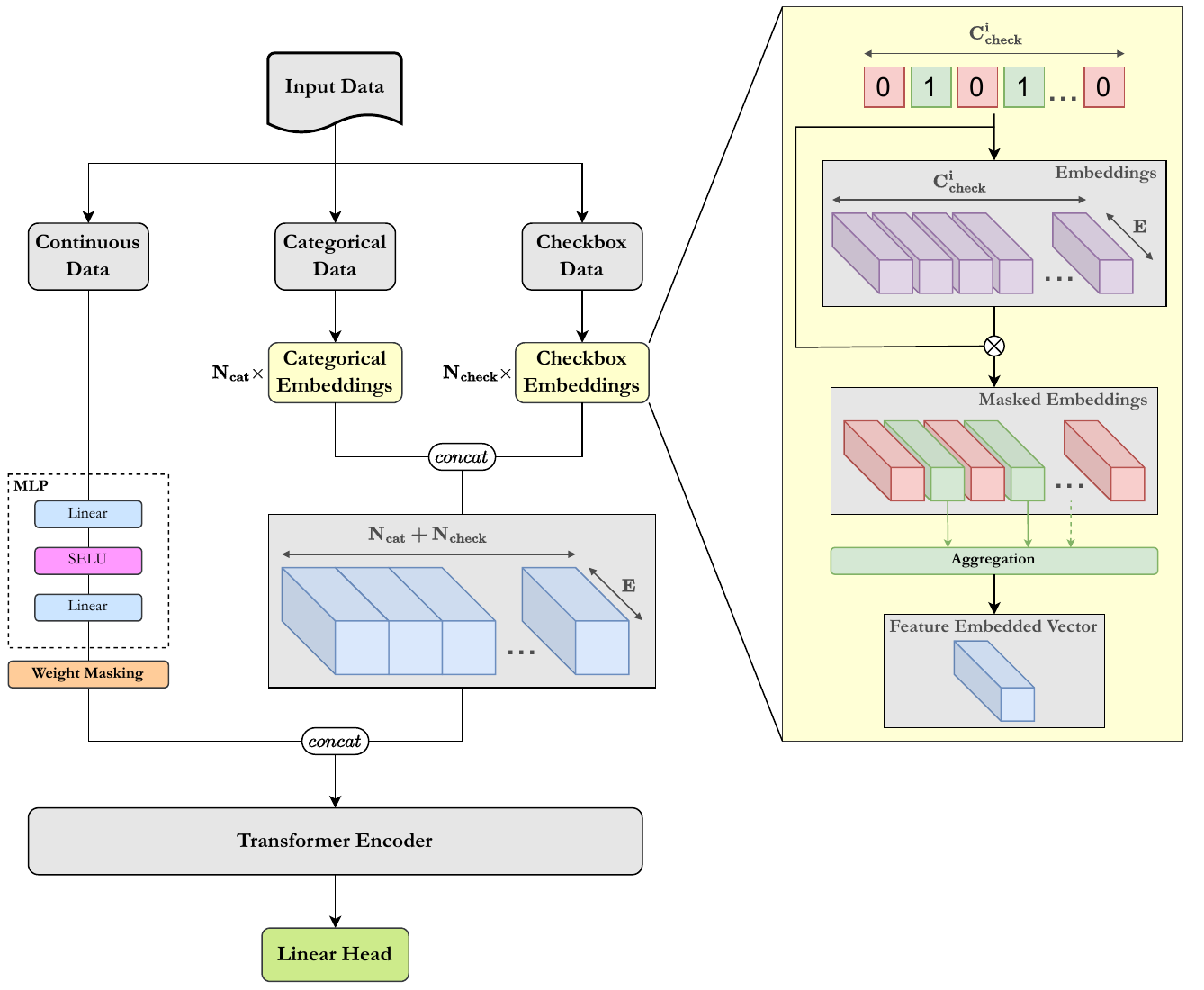}
\end{tabular}
\end{center}
\caption{The proposed TRACE model. The model supports three types of input data: continuous, categorical, and ``checkbox'' features. Continuous data are processed through a two-layer MLP with a weight masking mechanism on its outputs. Categorical and ``checkbox'' data are embedded through their respective embedding layers. All embeddings are concatenated and fed into a transformer encoder block followed by a linear head.
}
\label{fig:proposed}
\end{figure*}

\section{Methods}

\subsection{Non-negative neural network}
Inspired by \cite{Rieckmann2022}, we develop an architecture that features a three-layer non-negative MLP (nnMLP), shown in Figure~\ref{fig:nnmlp}. This design ensures that, during training, weights are constrained to be non-negative. In this context, exposures could either contribute positively to the risk outcome (when $w>0$) or have no effect (when $w=0$). Following the baseline, the biases of the two hidden layers are also constrained to be negative, allowing only sufficiently large weights to pass through the activation function, thereby positively affecting the final outcome. However, the bias term of the output layer is left unconstrained, providing additional flexibility in the model. Given the non-negativity of the network’s weights and the utilization of the sigmoid function as the output layer’s activation function, the unconstrained bias provides a baseline risk calculation. Finally, the weight parameters connecting the second hidden layer and the output layer are made learnable. Consequently, the output logit vector is characterized as a weighted sum of the second hidden layer’s activation outputs, rather than simply aggregating them.

Let $i, j, k$ denote the indices of the nodes in the input layer, the first hidden layer, and the second hidden layer, respectively, $X_{\{1...i\}}$ denote the input data features, $Z_{\{1...j\}}^1$, $Z_{\{1...k\}}^2$ the activation outputs of the first and second hidden layer, respectively. Letting $W, b$ denote the learnable weight parameters and biases of each layer and superscripts $^{(+)}, ^{(-)}$ the corresponding non-negativity and non-positivity constraints, respectively, equations (\ref{eq:mlp-z1}) - (\ref{eq:mlp-y}) describe the operations behind the proposed baseline architecture: 

\begin{gather}
	Z_j^1 = ReLU\left(\sum_{i} \left(W_{i,j}^{1(+)} \cdot X_i\right) + b_{j}^{1(-)} \right) \label{eq:mlp-z1},\\[10pt]
	Z_k^2 = ReLU\left(\sum_{j} \left(W_{j,k}^{2(+)} \cdot Z_j^1\right) + b_{k}^{2(-)} \right) \label{eq:mlp-z2},\\[10pt]
	Y = \sum_{k} \left(W_{k,1}^{3(+)} \cdot Z_k^2\right) + b^{3}.\label{eq:mlp-y}
\end{gather}

It is worth mentioning that $ReLU$ is employed as the non-linear activation function on both hidden layers. Additionally, the output logits $Y$, can be interpreted as probabilities of the positive class outcome, with values in range $[0, 1]$, through the application of the sigmoid activation function.

\subsection{TRACE model}
The architecture of the proposed Transformer-based Risk Assessment for Clinical Evaluation (TRACE) method is illustrated in Figure~\ref{fig:proposed}. It handles various data types encountered in clinical datasets, namely: numerical (continuous), categorical and ``checkbox'' data.  For continuous data, we employ a Multi-Layer Perceptron (MLP), consisting of two linear layers and a SELU activation function between them, to introduce non-linearity. The dimensions of the MLP output are $(B, N_{num}, E)$ where $B$ is the batch size, $N_{num}$ the number of numerical features available on the dataset and $E$ the selected embedding size. Regarding categorical data, we deploy standard categorical embeddings for each feature, ensuring their discrete nature is maintained, while creating the embedded tensor $(B, N_{cat}, E)$ where $N_{cat}$ represents the number of categorical features available in the dataset.

Additionally, we define ``checkbox'' data, as vectors receiving values $0$ or $1$, with ones representing the existence of the corresponding category and zeros its absence, but with the possibility of multiple positive categories on a single feature, thereby making it more complex to handle than standard one-hot encoded data. This type of data are processed with a custom embedding layer, where each checkbox feature is represented as a binary vector of size $C^i_{check}$ corresponding to the total number of possible categories within the feature for $i\in\{1,...,N_{check}\}$. Each category within the $i$-th checkbox feature, passes through a categorical embedding, resulting in a tensor of size $(B, C^i_{check}, E)$. Next, an element-wise multiplication is applied between the embedded tensor and the initial binary vector, serving as a mask, zeroing out all the embedded vectors of non-active categories. Subsequently, the active embeddings are aggregated allowing all valid categories to interact with each other, creating a single vector of size $E$, representing the combined information within the checkbox feature. This process is repeated $N_{check}$ times, resulting in the final embedding tensor $(B, N_{check}, E)$.

The concatenated embedded tensor $(B, N_{num}+N_{check}+N_{cat}, E)$ integrates all the diverse data representations into a unified feature space, serving as the input to the Transformer Encoder module. This module utilizes the multi-head self-attention mechanism, allowing the model to capture complex relationships and interactions across the entirety of the features. Finally, a linear head is employed for estimating the risk score.

The proposed architecture explicitly handles missing values. For continuous data, an element-wise weight masking is applied on the MLP outputs, masking out the weight vectors that correspond to missing numerical values in the input data. For categorical and checkbox data, a special embedding token, is defined for signaling missing values, effectively ignoring these entries.

\section{Experimental Evaluation}
\subsection{Datasets}

\paragraph*{Skin Cancer Risk Datasets} The first dataset considered in this work is dedicated to first primary melanoma classification, collected in the context of the iToBoS H2020 EU research project \cite{itobos_protocol}. We will refer to it as the \textbf{S}urvey-\textbf{B}ased \textbf{M}elanoma \textbf{S}tudy (SBMS) dataset. At the time this study was conducted, the dataset comprised a total of 415 patient records, with a ratio of 68.7\% positive melanoma diagnoses, to 31.3\% negative ones. The dataset consists of 29 features, four of which contain numerical values, two checkbox-type, while the rest of them are categorical. Participants provided informed consent for the data collection in written form. \added{
 PAD-UFES-20\cite{PAD} is a skin lesion dataset comprising 2,298 lesions with six lesion types and up to 23 clinical features. HIBA\cite{HIBA} includes 1,616 lesions with 10 lesion types and up to 11 clinical features, while HAM10000\cite{HAM} consists of 10,015 lesions with 7 lesion types and up to 5 clinical features. All datasets exhibit severe imbalance in their original multiclass formulations. Consequently, we adopted a binary classification approach by grouping the lesion types into two categories, \textit{malignant} and \textit{benign}. Under this approach, PAD-UFES-20 comprises 52.2\% benign and 47.8\% malignant lesions, HIBA comprises 53.5\% benign and 46.5\% malignant lesions, and HAM10000 comprises 80.5\% benign and 19.5\% malignant lesions. This binary perspective is particularly important for clinical applications, as the early detection of malignant lesions can significantly improve patient outcomes. Further details on the binary conversion process for each dataset are provided in the Appendix \ref{sec:app1}.}

\paragraph*{BRFSS Survey Dataset} This study utilizes data from the \textbf{B}ehavioral \textbf{R}isk \textbf{F}actor \textbf{S}urveillance \textbf{S}ystem survey conducted in 2014 and 2022\cite{CDCBRFSS}. Following \cite{Xie2019}, we utilize the 2014 dataset for \textit{type 2 diabetes} classification, with 139,266 respondents retained, of whom 21,587 have been diagnosed with type 2 diabetes. During preprocessing, 26 features related to this task were selected, while samples containing at least one missing value were ignored. To fully exploit the capabilities of the proposed architecture, we created another version of the same dataset, that retains respondents with missing values and numerical features are maintained as continuous values rather than being converted into categorical variables with arbitrary bins. This version includes 408,599 respondents in total, with 60,538 positive diagnoses. In the same manner, we utilize the 2022 dataset for \textit{heart attack/disease} classification, with 48 features selected and 440,111 respondents retained in total (39,751 diagnosed with heart attack/disease at least once). A version of this dataset that excludes instances with missing values, indicates a significant drop in the available samples, with just 1,745 respondents remaining. 

\subsection{Implementation Details}
We train our models in a workstation equipped with an NVIDIA RTX A5000 with 24GB of VRAM. Each experiment follows an 80\%-20\% training-validation split on the SBMS dataset and a 90\%-10\% ratio on PAD-UFES-20, HAM10000, HIBA and all BRFSS dataset variants. During training, we ensure that each batch maintains the positive-negative target label ratio of the corresponding dataset. We employ the Focal loss\cite{Lin2017ICCV} for training both the proposed non-negative MLP and TRACE models. The Focal loss is particularly effective for healthcare clinical datasets where the case group (disease presence) is significantly outnumbered by the control group (disease absence). Via the $\alpha$ hyperparameter of the Focal loss, one can adjust how much emphasis is given on hard examples, which in our context are the false negatives (instances where the model failed to detect a disease). Due to class imbalance, these cases are usually harder to identify during clinical risk assessment compared to false positives (instances where the model incorrectly identified a disease). The $\alpha$ value selection, relates to the degree of class imbalance in the dataset.  Finally, the use of the Focal loss implicitly achieves better alignment between the model's confidence and correctness \cite{NEURIPS2020_aeb7b30e}, a crucial property for risk estimation models where binary classification is used as a proxy task to assess the probability of having or developing a disease.

For the proposed non-negative MLP network, we employ a 5-fold stratified k-fold cross validation approach, to ensure that each split maintains the class distribution of the dataset. Unless stated otherwise, the model utilizes a hidden size of 64, while the hyperparameters of the Focal loss are $\gamma=2$ and $\alpha=\{0.5, 0.9, 0.6, 0.5, 0.5\}$ for SBMS, BRFSS~2022, PAD-UFES-20, HIBA and HAM10000 datasets, respectively. Non-negative MLPs are trained for $100$ epochs and optimized using the RMSprop optimization algorithm with a learning rate of $2\cdot10^{-4}$, a momentum value of $0.9$ and weight decay equal to $10^{-3}$.

Regarding the TRACE model, in order to mitigate the large contribution of the majority class that leads to overfitting, we follow a custom sampling pipeline during training, ensuring that each batch preserves the target class distribution found in each dataset. Training is performed for $100$ epochs using the Adam optimizer and a learning rate of $2\cdot10^{-4}$. The Focal loss function is considered with hyperparameters $\gamma=2$ and $\alpha=\{0.5, 0.8, 0.6, 0.5, 0.5\}$ for SBMS, BRFSS~2022, PAD-UFES-20, HIBA and HAM10000 datasets, respectively. We select the embedding size, the number of transformer encoder layers and the number of attention heads, based on the a hyperparameter search, accounting for the differences in the number and nature of features in each dataset. The MLP ratio for each transformer encoder layer was fixed at 4 across all experiments. 

The aggregation of checkbox embeddings is performed using the summation of the masked embedded vectors while the final representation of the Transformer Encoder, which is subsequently passed through the linear classifier, is derived by an average pooling along the output encoded vectors.

For categorical data, distinct values are used to represent each category, while the missing values are represented with zeros. In the case of non-negative MLPs, categorical data are transformed into one-hot encoded vectors, to meet the input requirements of the model. 

 For both architectures, the best model is determined by the highest F1-Score. Appendix \ref{sec:app-training} provides a more detailed analysis of the training hyperparameters.

\begin{table*}[htbp]
\caption{Comparison of clinical risk assessment algorithms on \added{SBMS, BRFSS-2022, PAD-UFES-20, HIBA and HAM10000 datasets, respectively.} Average performance metrics and their corresponding standard deviations are reported for each algorithm. Best values are shown in bold and second-best values are underlined.}
\label{tab:results_main}
\begin{center} 
\begin{tabular}{|c|l|ccccc|cc|}
\hline
& Method & Accuracy & F1-Score & Sensitivity & Specificity & BA & \#Params & \#Flops \\
\hline
\multirow{9}[2]{*}{\begin{sideways}SBMS\end{sideways}} & Logistic Regression & $75.9\% {\scriptstyle \: \pm 1.7\%}$ & $0.830 {\scriptstyle \: \pm 0.011}$ & $0.856 {\scriptstyle \: \pm 0.036}$ & $0.546 {\scriptstyle \: \pm 0.098}$ & $70.1\% {\scriptstyle \: \pm 3.6\%}$ & N/A & N/A\\
& XGBoost & $\underline{85.1\% {\scriptstyle \: \pm 2.5\%}}$ & $\underline{0.895 {\scriptstyle \: \pm 0.022}}$ & $\underline{0.957 {\scriptstyle \: \pm 0.020}}$ & $\underline{0.636 {\scriptstyle \: \pm 0.060}}$ & $\underline{79.7\% {\scriptstyle \: \pm 2.5\%}}$ & N/A & N/A\\

& nnMLP (Ours) & $82.4\% {\scriptstyle \: \pm 2.5\%}$ & $0.879 {\scriptstyle \: \pm 0.015}$ & $0.923 {\scriptstyle \: \pm 0.024}$ & $0.608 {\scriptstyle \: \pm 0.104}$ & $76.5\% {\scriptstyle \: \pm 4.5\%}$ & 9,665 & 14,040\\

& TabNet & $71.4\% {\scriptstyle \: \pm 4.2\%}$ & $0.827 {\scriptstyle \: \pm 0.017}$ & $\mathbf{0.986 {\scriptstyle \: \pm 0.108}}$ & $0.201 {\scriptstyle \: \pm 0.104}$ & $54.7\% {\scriptstyle \: \pm 8.5\%}$ & 698,180 & 1,256,780\\

& FT-Transformer & ${75.2\% {\scriptstyle \: \pm 4.4\%}}$ & ${0.846 {\scriptstyle \: \pm 0.021}}$ & ${0.980 {\scriptstyle \: \pm 0.019}}$ & ${0.246 {\scriptstyle \: \pm 0.167}}$ & ${61.3\% {\scriptstyle \: \pm 7.7\%}}$ & 472,882 & 31,568,308\\

& DANET & $77.2\% {\scriptstyle \: \pm 2.7\%}$ & $0.850 {\scriptstyle \: \pm 0.014}$ & $0.938 {\scriptstyle \: \pm 0.038}$ & $0.400 {\scriptstyle \: \pm 0.148}$ & $66.9\% {\scriptstyle \: \pm 5.9\%}$ & 942,074 & 963,058\\

& GANDALF & $82.4\% {\scriptstyle \: \pm 3.2\%}$ & $0.878 {\scriptstyle \: \pm 0.019}$ & $0.911 {\scriptstyle \: \pm 0.019}$ & $0.630 {\scriptstyle \: \pm 0.127}$ & $77.1\% {\scriptstyle \: \pm 5.7\%}$ & 564,000 & 560,108\\

& GATE & $79.1\% {\scriptstyle \: \pm 3.1\%}$ & $0.853 {\scriptstyle \: \pm 0.022}$ & $0.883 {\scriptstyle \: \pm 0.083}$ & $0.584 {\scriptstyle \: \pm 0.208}$ & $73.4\% {\scriptstyle \: \pm 7.0\%}$ & 651,806 & 602,612\\

& TRACE (Ours) & $\mathbf{86.0\% {\scriptstyle \: \pm 3.7\%}}$ & $\mathbf{0.900 {\scriptstyle \: \pm 0.027}}$ & $0.941 {\scriptstyle \: \pm 0.018}$ & $\mathbf{0.709 {\scriptstyle \: \pm 0.121}}$ & $\mathbf{82.5\% {\scriptstyle \: \pm 5.3\%}}$ & 285,953 & 7,497,753\\

\hline

\multirow{9}[2]{*}{\begin{sideways}BRFSS 2022\end{sideways}} & Logistic Regression & $\mathbf{91.1\% {\scriptstyle \: \pm 0.0\%}}$ & $0.193 {\scriptstyle \: \pm 0.009}$ & $0.117 {\scriptstyle \: \pm 0.007}$ & $\mathbf{0.990 {\scriptstyle \: \pm 0.001}}$ & $55.3\% {\scriptstyle \: \pm 0.2\%}$ & N/A & N/A\\
& XGBoost & $\mathbf{91.1\% {\scriptstyle \: \pm 0.0\%}}$ & $0.202 {\scriptstyle \: \pm 0.006}$ & ${0.124 {\scriptstyle \: \pm 0.004}}$ & $\mathbf{0.990 {\scriptstyle \: \pm 0.001}}$ & $55.7\% {\scriptstyle \: \pm 0.2\%}$ & N/A & N/A\\

& nnMLP (Ours) & $81.2\% {\scriptstyle \: \pm 0.5\%}$ & $0.399 {\scriptstyle \: \pm 0.003}$ & $\mathbf{0.692 {\scriptstyle \: \pm 0.012}}$ & $0.824 {\scriptstyle \: \pm 0.006}$ & $\mathbf{75.8\% {\scriptstyle \: \pm 0.3\%}}$ & 17,409 & 17,312\\

& TabNet & $85.9\% {\scriptstyle \: \pm 0.8\%}$ & $\underline{0.411 {\scriptstyle \: \pm 0.003}}$ & $0.547 {\scriptstyle \: \pm 0.037}$ & $\underline{0.890 {\scriptstyle \: \pm 0.012}}$ & $71.9\% {\scriptstyle \: \pm 1.2\%}$ & 494,744 & 899,039\\

& FT-Transformer & $ {82.8\% {\scriptstyle \: \pm 6.3\%}}$ & ${0.394 {\scriptstyle \: \pm 0.026}}$ & $\underline{0.606 {\scriptstyle \: \pm 0.152}}$ & ${0.850 {\scriptstyle \: \pm 0.084}}$ & ${72.8\% {\scriptstyle \: \pm 3.5\%}}$ & 4,366,810 & 241,822,030\\

& DANET & $84.3\% {\scriptstyle \: \pm 3.7\%}$ & $0.404 {\scriptstyle \: \pm 0.019}$ & $0.583 {\scriptstyle \: \pm 0.096}$ & $0.870 {\scriptstyle \: \pm 0.050}$ & $72.6\% {\scriptstyle \: \pm 2.5\%}$ & 4,543,161 & 4,623,646\\

& GANDALF & $\underline{87.0\% {\scriptstyle \: \pm 3.8\%}}$ & $0.375 {\scriptstyle \: \pm 0.047}$ & $0.449 {\scriptstyle \: \pm 0.183}$ & $0.912 {\scriptstyle \: \pm 0.060}$ & $68.0\% {\scriptstyle \: \pm 6.2\%}$ & 1,044,538 & 1,037,556\\

& GATE & $84.1\% {\scriptstyle \: \pm 2.3\%}$ & $0.396 {\scriptstyle \: \pm 0.037}$ & $0.573 {\scriptstyle \: \pm 0.019}$ & $0.868 {\scriptstyle \: \pm 0.023}$ & $72.1\% {\scriptstyle \: \pm 1.9\%}$ & 588,704 & 548,986\\

& TRACE (Ours) & ${85.7\% {\scriptstyle \: \pm 0.6\%}}$ & $\mathbf{{0.422 {\scriptstyle \: \pm 0.006}}}$ & $0.577 {\scriptstyle \: \pm 0.012}$ & ${0.885 {\scriptstyle \: \pm 0.007}}$ & $\underline{73.1\% {\scriptstyle \: \pm 0.3\%}}$ & 328,449 & 10,410,368\\
\hline
\multirow{9}[2]{*}{\begin{sideways}PAD-UFES-20\end{sideways}} & Logistic Regression & $83.7\% {\scriptstyle \: \pm 2.1\%}$ & $0.842 {\scriptstyle \: \pm 0.017}$ & $0.905 {\scriptstyle \: \pm 0.010}$ & $0.775 {\scriptstyle \: \pm 0.042}$ & $84.0\% {\scriptstyle \: \pm 2.0\%}$ & N/A & N/A\\
& XGBoost & $\mathbf{92.4\% {\scriptstyle \: \pm 1.8\%}}$ & $\mathbf{0.924 {\scriptstyle \: \pm 0.016}}$ & $0.960 {\scriptstyle \: \pm 0.019}$ & $\underline{0.892 {\scriptstyle \: \pm 0.036}}$ & $\mathbf{92.6\% {\scriptstyle \: \pm 1.7\%}}$ & N/A & N/A\\

& nnMLP (Ours) & $90.5\% {\scriptstyle \: \pm 1.6\%}$ & $0.908 {\scriptstyle \: \pm 0.013}$ & $0.973 {\scriptstyle \: \pm 0.014}$ & $0.843 {\scriptstyle \: \pm 0.044}$ & $90.8\% {\scriptstyle \: \pm 1.5\%}$ & 8,129 & 8,032\\

& TabNet & $82.6\% {\scriptstyle \: \pm 0.8\%}$ & $0.844 {\scriptstyle \: \pm 0.006}$ & $\mathbf{0.980 {\scriptstyle \: \pm 0.028}}$ & $0.685 {\scriptstyle \: \pm 0.033}$ & $83.2\% {\scriptstyle \: \pm 0.7\%}$ & 392,668 & 604,839\\

& FT-Transformer & ${82.9\% {\scriptstyle \: \pm 0.8\%}}$ & ${0.842 {\scriptstyle \: \pm 0.007}}$ & ${0.955 {\scriptstyle \: \pm 0.039}}$ & ${0.715 {\scriptstyle \: \pm 0.044}}$ & ${83.5\% {\scriptstyle \: \pm 0.7\%}}$ & 2,771,698 & 69,784,884\\

& DANET & $85.6\% {\scriptstyle \: \pm 3.1\%}$ & $0.864 {\scriptstyle \: \pm 0.025}$ & $0.949 {\scriptstyle \: \pm 0.026}$ & $0.770 {\scriptstyle \: \pm 0.072}$ & $86.0\% {\scriptstyle \: \pm 3.0\%}$ & 1,684,853 & 1,740,686\\

& GANDALF & $80.8\% {\scriptstyle \: \pm 2.1\%}$ & $0.802 {\scriptstyle \: \pm 0.036}$ & $0.827 {\scriptstyle \: \pm 0.111}$ & $0.790 {\scriptstyle \: \pm 0.085}$ & $80.9\% {\scriptstyle \: \pm 2.3\%}$ & 169,554 & 167,226\\

& GATE & $85.1\% {\scriptstyle \: \pm 3.2\%}$ & $0.863 {\scriptstyle \: \pm 0.024}$ & $\underline{0.975 {\scriptstyle \: \pm 0.024}}$ & $0.736 {\scriptstyle \: \pm 0.079}$ & $85.6\% {\scriptstyle \: \pm 3.0\%}$ & 505,692 & 347,768\\

& TRACE (Ours) & $\underline{92.3\% {\scriptstyle \: \pm 2.8\%}}$ & $\underline{0.921 {\scriptstyle \: \pm 0.028}}$ & $0.934 {\scriptstyle \: \pm 0.021}$ & $\mathbf{0.913 {\scriptstyle \: \pm 0.040}}$ & $\underline{92.4\% {\scriptstyle \: \pm 2.8\%}}$ & 674,945 & 14,756,480\\

\hline

\multirow{9}[2]{*}{\begin{sideways}HIBA\end{sideways}} & Logistic Regression & $86.4\% {\scriptstyle \: \pm 2.9\%}$ & $0.865 {\scriptstyle \: \pm 0.028}$ & $0.941 {\scriptstyle \: \pm 0.038}$ & $0.798 {\scriptstyle \: \pm 0.041}$ & $87.0\% {\scriptstyle \: \pm 2.8\%}$ & N/A & N/A\\
& XGBoost & $87.3\% {\scriptstyle \: \pm 2.4\%}$ & $0.865 {\scriptstyle \: \pm 0.026}$ & ${0.869 {\scriptstyle \: \pm 0.042}}$ & $0.876 {\scriptstyle \: \pm 0.042}$ & $87.3\% {\scriptstyle \: \pm 2.4\%}$ & N/A & N/A\\

& nnMLP (Ours) & $\underline{91.1\% {\scriptstyle \: \pm 1.6\%}}$ & $\underline{0.907 {\scriptstyle \: \pm 0.014}}$ & $0.936 {\scriptstyle \: \pm 0.035}$ & $\mathbf{0.890 {\scriptstyle \: \pm 0.048}}$ & $\underline{91.3\% {\scriptstyle \: \pm 1.4\%}}$ & 4,417 & 4,320\\

& TabNet & $83.7\% {\scriptstyle \: \pm 1.8\%}$ & $0.844 {\scriptstyle \: \pm 0.021}$ & $0.952 {\scriptstyle \: \pm 0.050}$ & $0.738 {\scriptstyle \: \pm 0.026}$ & $84.5\% {\scriptstyle \: \pm 1.9\%}$ & 789,940 & 1,215,726\\

& FT-Transformer & $ {84.6\% {\scriptstyle \: \pm 1.0\%}}$ & ${0.842 {\scriptstyle \: \pm 0.020}}$ & $0.899 {\scriptstyle \: \pm 0.095}$ & ${0.800 {\scriptstyle \: \pm 0.076}}$ & ${84.9\% {\scriptstyle \: \pm 1.4\%}}$ & 1,843,882 & 22,671,602\\

& DANET & $84.2\% {\scriptstyle \: \pm 0.9\%}$ & $0.843 {\scriptstyle \: \pm 0.015}$ & $0.923 {\scriptstyle \: \pm 0.072}$ & $0.773 {\scriptstyle \: \pm 0.057}$ & $84.8\% {\scriptstyle \: \pm 1.1\%}$ & 1,221,654 & 1,279,638\\

& GANDALF & $86.0\% {\scriptstyle \: \pm 0.7\%}$ & $0.855 {\scriptstyle \: \pm 0.013}$ & $0.893 {\scriptstyle \: \pm 0.057}$ & $0.833 {\scriptstyle \: \pm 0.043}$ & $86.3\% {\scriptstyle \: \pm 1.0\%}$ & 7,282 & 6,970\\

& GATE & $85.9\% {\scriptstyle \: \pm 1.0\%}$ & $0.864 {\scriptstyle \: \pm 0.008}$ & $\mathbf{0.966 {\scriptstyle \: \pm 0.020}}$ & $0.768 {\scriptstyle \: \pm 0.029}$ & $86.7\% {\scriptstyle \: \pm 0.9\%}$ & 36,722 & 26,954\\

& TRACE (Ours) & ${\mathbf{91.4\% {\scriptstyle \: \pm 1.0\%}}}$ & $\mathbf{{0.911 {\scriptstyle \: \pm 0.008}}}$ & $\underline{0.952 {\scriptstyle \: \pm 0.015}}$ & ${\underline{0.880 {\scriptstyle \: \pm 0.029}}}$ & $\mathbf{91.6\% {\scriptstyle \: \pm 0.8\%}}$ & 633,089 & 7,267,712\\

\hline

\multirow{9}[2]{*}{\begin{sideways}HAM10000\end{sideways}} & Logistic Regression & ${81.5\% {\scriptstyle \: \pm 0.6\%}}$ & $0.407 {\scriptstyle \: \pm 0.046}$ & $0.328 {\scriptstyle \: \pm 0.051}$ & $\mathbf{0.933 {\scriptstyle \: \pm 0.006}}$ & $63.1\% {\scriptstyle \: \pm 2.3\%}$ & N/A & N/A\\
& XGBoost & $\mathbf{83.2\% {\scriptstyle \: \pm 0.6\%}}$ & $0.531 {\scriptstyle \: \pm 0.029}$ & ${0.491 {\scriptstyle \: \pm 0.042}}$ & $\underline{0.914 {\scriptstyle \: \pm 0.007}}$ & $70.3\% {\scriptstyle \: \pm 1.9\%}$ & N/A & N/A\\

& nnMLP (Ours) & $82.6\% {\scriptstyle \: \pm 1.8\%}$ & $\underline{0.619 {\scriptstyle \: \pm 0.024}}$ & ${0.723 {\scriptstyle \: \pm 0.044}}$ & $0.851 {\scriptstyle \: \pm 0.027}$ & ${78.7\% {\scriptstyle \: \pm 1.8\%}}$ & 3,713 & 3,616\\

& TabNet & $78.8\% {\scriptstyle \: \pm 4.9\%}$ & ${0.538 {\scriptstyle \: \pm 0.067}}$ & $0.660 {\scriptstyle \: \pm 0.241}$ & ${0.819{\scriptstyle \: \pm 0.113}}$ & $74.0\% {\scriptstyle \: \pm 7.0\%}$ & 647,843 & 968,939\\

& FT-Transformer & $ {79.9\% {\scriptstyle \: \pm 0.3\%}}$ & ${0.558 {\scriptstyle \: \pm 0.091}}$ & ${0.684 {\scriptstyle \: \pm 0.216}}$ & ${0.827 {\scriptstyle \: \pm 0.051}}$ & ${75.6\% {\scriptstyle \: \pm 8.3\%}}$ & 462,010 & 2,841,730\\

& DANET & $74.8\% {\scriptstyle \: \pm 5.2\%}$ & $0.586 {\scriptstyle \: \pm 0.030}$ & $\underline{0.907 {\scriptstyle \: \pm 0.072}}$ & $0.709 {\scriptstyle \: \pm 0.082}$ & $\underline{80.8\% {\scriptstyle \: \pm 1.0\%}}$ & 431,680 & 455,036\\

& GANDALF & ${80.9\% {\scriptstyle \: \pm 1.6\%}}$ & $0.438 {\scriptstyle \: \pm 0.098}$ & $0.404 {\scriptstyle \: \pm 0.163}$ & $0.908 {\scriptstyle \: \pm 0.049}$ & $65.6\% {\scriptstyle \: \pm 5.9\%}$ & 3,011 & 2,740\\

& GATE & $76.2\% {\scriptstyle \: \pm 4.4\%}$ & $0.605 {\scriptstyle \: \pm 0.033}$ & $\mathbf{0.928 {\scriptstyle \: \pm 0.047}}$ & $0.722 {\scriptstyle \: \pm 0.066}$ & $\mathbf{82.5\% {\scriptstyle \: \pm 1.3\%}}$ & 1,033,629 & 523,370\\

& TRACE (Ours) & $\underline{82.7\% {\scriptstyle \: \pm 1.5\%}}$ & $\mathbf{{0.635 {\scriptstyle \: \pm 0.011}}}$ & $0.774 {\scriptstyle \: \pm 0.053}$ & ${0.839 {\scriptstyle \: \pm 0.030}}$ & ${80.7\% {\scriptstyle \: \pm 1.3\%}}$ & 160,065 & 918,464\\
\hline

\end{tabular}
\end{center}
\end{table*}

\begin{table*}[htbp]
\caption{Comparison of clinical risk assessment algorithms on the mutliclass classification setup of PAD-UFES-20, HIBA and HAM10000 datasets, respectively. Average performance metrics and their corresponding standard deviations are reported for each algorithm. Best values are shown in bold and second-best values are underlined.}
\label{tab:results_multiclass}
\begin{center} 
\begin{tabular}{|c|l|ccc|c|}
\hline
& Method & Accuracy & F1-Score (macro) & F1-Score (weighted) & \#Params \\
\hline
\multirow{9}[2]{*}{\begin{sideways}PAD-UFES-20\end{sideways}} & Logistic Regression & $73.4\% {\scriptstyle \: \pm 2.8\%}$ & $0.577 {\scriptstyle \: \pm 0.020}$ & $0.700 {\scriptstyle \: \pm 0.023}$ & N/A \\
& XGBoost & $\underline{81.6\% {\scriptstyle \: \pm 3.2\%}}$ & $\underline{0.746 {\scriptstyle \: \pm 0.067}}$ & $\mathbf{0.813 {\scriptstyle \: \pm 0.031}}$ & N/A\\

& nnMLP (Ours) & $79.4\% {\scriptstyle \: \pm 2.2\%}$ & $0.721 {\scriptstyle \: \pm 0.028}$ & $0.788 {\scriptstyle \: \pm 0.021}$  & 8,294\\

& TabNet & $62.3\% {\scriptstyle \: \pm 1.2\%}$ & $0.319 {\scriptstyle \: \pm 0.048}$ & $0.541 {\scriptstyle \: \pm 0.027}$ & 392,917\\

& FT-Transformer & ${75.1\% {\scriptstyle \: \pm 1.4\%}}$ & ${0.606 {\scriptstyle \: \pm 0.052}}$ & ${0.734 {\scriptstyle \: \pm 0.020}}$ & 2,772,086\\

& DANET & $75.9\% {\scriptstyle \: \pm 0.6\%}$ & $0.625 {\scriptstyle \: \pm 0.048}$ & $0.738 {\scriptstyle \: \pm 0.009}$ & 1,685,106\\

& GANDALF & $78.7\% {\scriptstyle \: \pm 3.0\%}$ & $0.625 {\scriptstyle \: \pm 0.048}$ & $0.738 {\scriptstyle \: \pm 0.009}$ & 169,791\\

& GATE & $75.6\% {\scriptstyle \: \pm 3.9\%}$ & $0.561 {\scriptstyle \: \pm 0.121}$ & $0.730 {\scriptstyle \: \pm 0.061}$ & 505,817\\

& TRACE (Ours) & $\mathbf{83.2\% {\scriptstyle \: \pm 2.1\%}}$ & $\mathbf{0.783 {\scriptstyle \: \pm 0.121}}$ & $\underline{0.811 {\scriptstyle \: \pm 0.062}}$ & 675,590\\

\hline

\multirow{9}[2]{*}{\begin{sideways}HIBA\end{sideways}} & Logistic Regression & $57.8\% {\scriptstyle \: \pm 2.9\%}$ & $0.302 {\scriptstyle \: \pm 0.019}$ & $0.543 {\scriptstyle \: \pm 0.021}$ & N/A \\
& XGBoost & $\mathbf{69.9\% {\scriptstyle \: \pm 5.2\%}}$ & $\underline{0.529 {\scriptstyle \: \pm 0.069}}$ & $\mathbf{0.694 {\scriptstyle \: \pm 0.046}}$ & N/A\\

& nnMLP (Ours) & $64.5\% {\scriptstyle \: \pm 2.8\%}$ & $0.433 {\scriptstyle \: \pm 0.032}$ & $0.620 {\scriptstyle \: \pm 0.038}$  & 4,714\\

& TabNet & $54.8\% {\scriptstyle \: \pm 3.9\%}$ & $0.176 {\scriptstyle \: \pm 0.040}$ & $0.440 {\scriptstyle \: \pm 0.059}$ & 790,452\\

& FT-Transformer & ${65.6\% {\scriptstyle \: \pm 1.7\%}}$ & ${0.442 {\scriptstyle \: \pm 0.083}}$ & ${0.615 {\scriptstyle \: \pm 0.034}}$ & 1,844,914\\

& DANET & $64.8\% {\scriptstyle \: \pm 2.5\%}$ & $0.447 {\scriptstyle \: \pm 0.105}$ & $0.610 {\scriptstyle \: \pm 0.047}$ & 1,222,174\\

& GANDALF & $67.5\% {\scriptstyle \: \pm 1.8\%}$ & $0.526 {\scriptstyle \: \pm 0.035}$ & $0.653 {\scriptstyle \: \pm 0.021}$ & 7,490\\

& GATE & $64.4\% {\scriptstyle \: \pm 2.0\%}$ & $0.476 {\scriptstyle \: \pm 0.031}$ & $0.608 {\scriptstyle \: \pm 0.015}$ & 36,794\\

& TRACE (Ours) & $\underline{67.9\% {\scriptstyle \: \pm 2.2\%}}$ & $\mathbf{0.556 {\scriptstyle \: \pm 0.010}}$ & $\underline{0.675 {\scriptstyle \: \pm 0.017}}$ & 634,250\\

\hline

\multirow{9}[2]{*}{\begin{sideways}HAM10000\end{sideways}} & Logistic Regression & $67.6\% {\scriptstyle \: \pm 0.9\%}$ & $0.182 {\scriptstyle \: \pm 0.013}$ & $0.615 {\scriptstyle \: \pm 0.009}$ & N/A \\
& XGBoost & $\mathbf{74.2\% {\scriptstyle \: \pm 1.0\%}}$ & $\mathbf{0.428 {\scriptstyle \: \pm 0.035}}$ & $\underline{0.718 {\scriptstyle \: \pm 0.011}}$  & N/A\\

& nnMLP (Ours) & $70.0\% {\scriptstyle \: \pm 0.8\%}$ & $0.298 {\scriptstyle \: \pm 0.011}$ & $0.673 {\scriptstyle \: \pm 0.008}$  & 3,911\\

& TabNet & $71.4\% {\scriptstyle \: \pm 0.2\%}$ & $0.325 {\scriptstyle \: \pm 0.031}$ & ${0.673 {\scriptstyle \: \pm 0.018}}$ & 648,163\\

& FT-Transformer & ${70.3\% {\scriptstyle \: \pm 0.2\%}}$ & ${0.280 {\scriptstyle \: \pm 0.047}}$ & ${0.653 {\scriptstyle \: \pm 0.011}}$ & 462,335\\

& DANET & $71.9\% {\scriptstyle \: \pm 0.5\%}$ & $0.371 {\scriptstyle \: \pm 0.011}$ & $0.688 {\scriptstyle \: \pm 0.010}$ & 432,005\\

& GANDALF & $71.4\% {\scriptstyle \: \pm 0.8\%}$ & $0.357 {\scriptstyle \: \pm 0.039}$ & $0.678 {\scriptstyle \: \pm 0.014}$ & 3,096\\

& GATE & $70.6\% {\scriptstyle \: \pm 0.6\%}$ & $0.216 {\scriptstyle \: \pm 0.032}$ & $0.634 {\scriptstyle \: \pm 0.014}$ & 1,033,794\\

& TRACE (Ours) & $\underline{73.5\% {\scriptstyle \: \pm 1.8\%}}$ & $\underline{0.425 {\scriptstyle \: \pm 0.020}}$ & $\mathbf{0.720 {\scriptstyle \: \pm 0.016}}$ & 160,455\\
\hline

\end{tabular}
\end{center}
\end{table*}

\subsection{Evaluation}
Evaluation is performed considering five metrics: Accuracy, F1-Score, Sensitivity and Specificity and Balanced Accuracy (BA). Accuracy is a straightforward metric, defined as the ratio of correctly predicted examples, both true negatives and true positives, to the total number of instances evaluated. F1-Score is the harmonic mean of Precision and Recall of the positive class to assess the model's performance on imbalanced datasets. Sensitivity (or Recall), and Specificity are widely used in clinical tasks, illustrating the model's ability to correctly identify positive and negative instances, respectively, while BA represents their average score.

Table \ref{tab:results_main} compares the proposed TRACE method with various baselines on the skin cancer and BRFSS22 datasets. In particular, Logistic Regression and XGBoost, two ML methods widely used in clinical risk assessment, are considered as baselines, along with the proposed non-negative MLP (nnMLP). Additionally, we evaluate TRACE alognside several deep learning models for tabular data, including the FT-Transformer~\cite{Gorishniy2021RevisitingDL}, TabNet~\cite{TabNet}, DANET~\cite{DANET}, GANDALF~\cite{GANDALF} and GATE~\cite{GATE}. Additional details about the training strategy followed for the selected baselines are provided in the Appendix~\ref{sec:app-training}.

Table~\ref{tab:results_main} presents the results from five runs on the five datasets regarding the tasks of binary classification, each with different random seeds, while keeping the training/validation splits consistent across all baselines considered in this study. Specifically, for the SBMS dataset our model outperforms the baselines in terms of Accuracy, F1-Score, Specificity and Balanced Accuracy (BA) metrics, with fewer trainable parameters compared to the state-of-the-art methods. 

Regarding the BRFSS22 dataset, it introduces a more challenging task due to its highly imbalanced class distribution. In such scenarios, higher Accuracy alone may not reflect higher generalization, thus metrics like Accuracy and Specificity can be misleading, as observed with both ML algorithms. Instead, F1-Score and BA metrics offer a more representative overall assessment of the model's performance. Our proposed model significantly outperforms the baselines regarding the F1-Score and achieves the second-best BA, just behind the proposed nnMLP. These results demonstrate that our method provides robust and balanced performance across all scenarios. Notably, nnMLP achieves the best performance in terms of Balanced Accuracy, while requiring significantly fewer trainable parameters than any other baseline.

Across the remaining three skin cancer datasets, our proposed models consistently demonstrate strong and reliable performance across key evaluation metrics.
For the PAD-UFES-20 dataset, TRACE achieves the second-best performance in both F1-Score and Balanced Accuracy, with results that are very close to the top-performing model. Notably, nnMLP and TRACE maintains a significant lead over all other baselines.
In the case of the HIBA dataset, the proposed methods dominate most of the quantitative benchmarks. TRACE outperforms all other models by achieving the highest values for both Balanced Accuracy and F1-Score. Additionally, the nnMLP model delivers very strong results, especially considering its minimal parameter count and computational cost (FLOPs). This underscores the efficiency and practicality of the nnMLP method for real-world applications with limited resources.
Finally, on the HAM10000 dataset which has high class imbalance and the fewest attributes of all the considered datasets, our models continue to perform competitively. TRACE secures the highest F1-Score across all baselines, demonstrating its robustness in handling imbalanced data. While GATE achieves a slightly higher Balanced Accuracy, TRACE stands out by delivering the best F1-Score, emphasizing its effectiveness in capturing both precision and recall,particularly important in this highly imbalanced classification scenario.

Table \ref{tab:results_multiclass} presents the results of the multiclass classification task, conducted under the same experimental setup as the binary classification experiments. The datasets used for this evaluation are PAD-UFES-20, HIBA, and HAM10000, each offering different characteristics and challenges. To assess model performance, we considered two key metrics: Accuracy and F1-Score. The F1-Score is further broken down into two variants: macro and weighted. The macro F1-Score calculates the unweighted mean of F1 scores for all classes, treating each class equally regardless of its frequency. In contrast, the weighted F1-Score takes class imbalance into account by assigning higher weight to more frequent classes.
On the PAD-UFES-20 dataset, our proposed model outperforms all baselines in both Accuracy and macro F1-Score, highlighting its effectiveness in treating all classes fairly, including those with fewer examples. Although XGBoost slightly surpasses our model in weighted F1-Score, our results remain competitive and show strong performance across the board.
In the HIBA dataset, our model indicates solid performance by achieving the highest macro F1-Score. Additionally, it ranks second in both Accuracy and weighted F1-Score, while maintaining a notable margin over all other baseline methods, further showcasing the model’s consistency and generalization ability.
Finally, for the HAM10000 dataset our approach achieves the highest weighted F1-Score. While it ranks second in Accuracy and macro F1-Score, the overall performance remains among the top across all metrics.
\begin{table}[htbp]
\caption{Comparison of the proposed TRACE model, considering different $\mathbf{\alpha}$ values of the Focal loss, with \cite{Xie2019} on the BRFSS~2014 dataset. Best values are shown in bold and second-best values are underlined.}
\label{tab:results_brfss14}
\vspace{-0.5cm}
\begin{center} 
\resizebox{\linewidth}{!}{%
\begin{tabular}{|l|cccc|}
\hline
Method & Accuracy & Sensitivity & Specificity & BA \\
\hline
Neural network\cite{Xie2019} & ${82.4\%}$ & $0.378$ & ${0.902}$ & $64.0\%$\\
\hline
TRACE (Ours), $\alpha=0.5$ & $\mathbf{84.1\%}$ & $0.332$ & $\mathbf{0.934}$ & $63.3\%$\\
\hline
TRACE (Ours), $\alpha=0.6$ & $81.8\%$ & $0.500$ & $0.877$ & $68.9\%$\\
\hline
TRACE (Ours), $\alpha=0.7$ & $79.4\%$ & ${0.595}$ & $0.831$ & ${71.3\%}$\\
\hline
TRACE (Ours), $\alpha=0.8$ & $77.2\%$ & $\mathbf{0.670}$ & $0.790$ & $\mathbf{73.0\%}$\\
\hline
\end{tabular}
}
\end{center}
\end{table}

Table~\ref{tab:results_brfss14} compares the performance of the proposed TRACE model, with the Neural Network from \cite{Xie2019} that outperformed all the baselines for the task of type 2 diabetes risk prediction on BRFSS~2014 dataset.  The dataset is pre-processed based on the guidelines by \cite{Xie2019}. Our model is evaluated using different values of the Focal Loss hyperparameter $\alpha$ and as shown, it consistently outperforms the neural network by \cite{Xie2019} in several key metrics. More specifically, for higher $\alpha$ values, our model achieves the best performance in terms of Sensitivity and Balanced Accuracy metrics, reflecting its robustness in identifying positive cases while maintaining a balanced performance between the classes.

\begin{figure*}[t!]
\begin{center}
\begin{tabular}{cc} 
\hspace{1.3cm} \includegraphics[height=6.9cm]{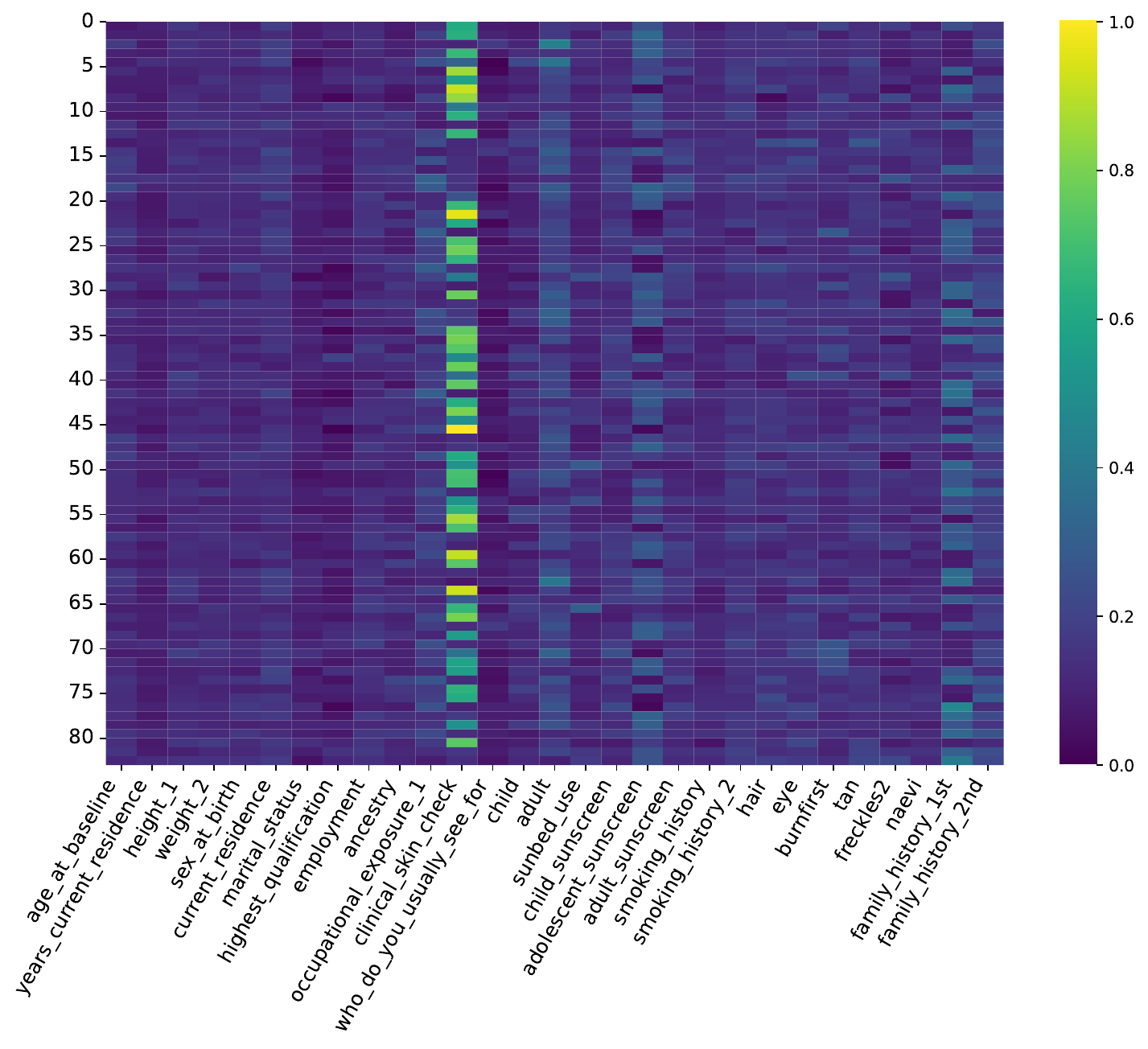} &
\hspace{0.7cm} \includegraphics[height=6.9cm,trim={0 -2cm 0 0}]{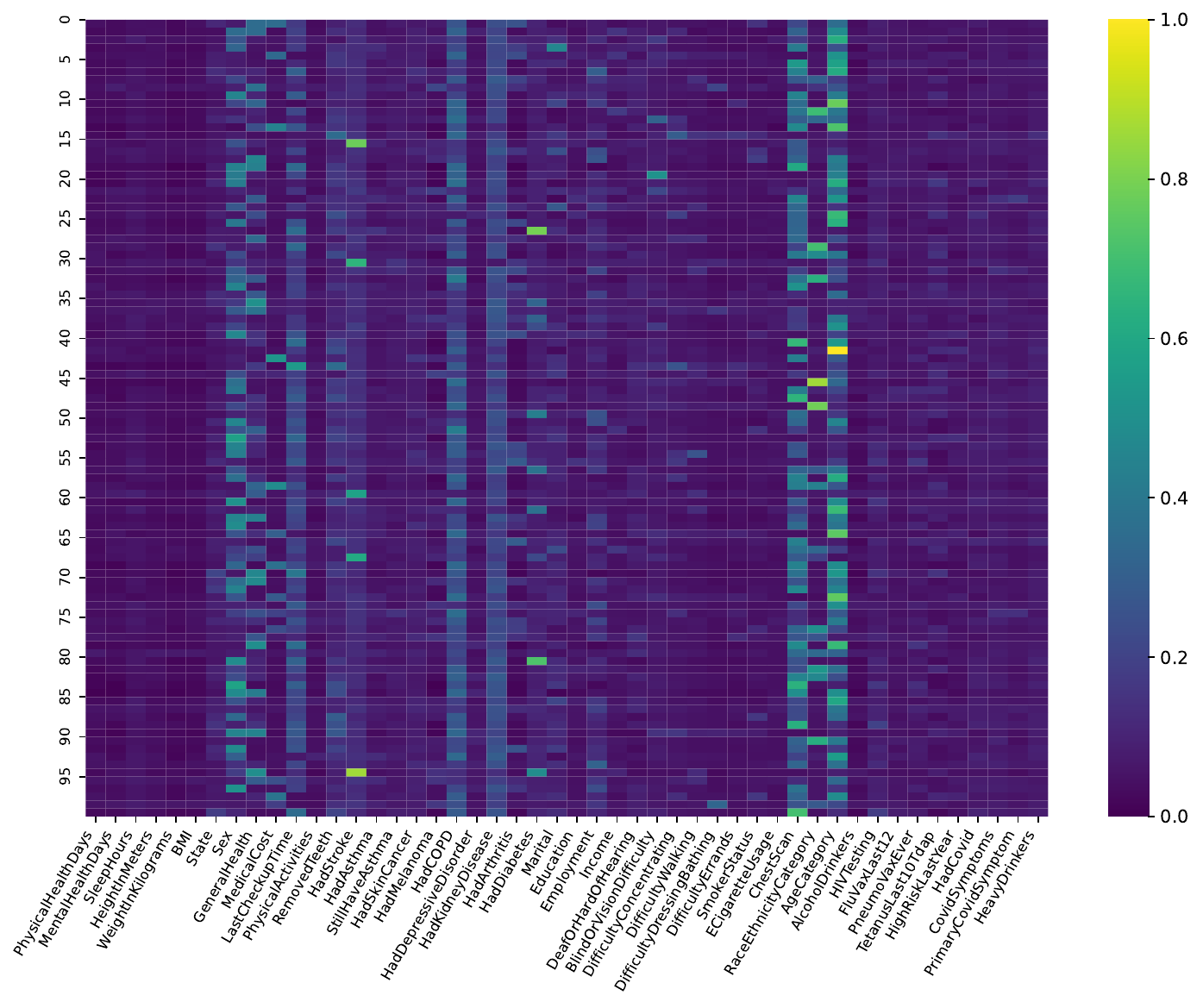}\\
\includegraphics[height=6.9cm]{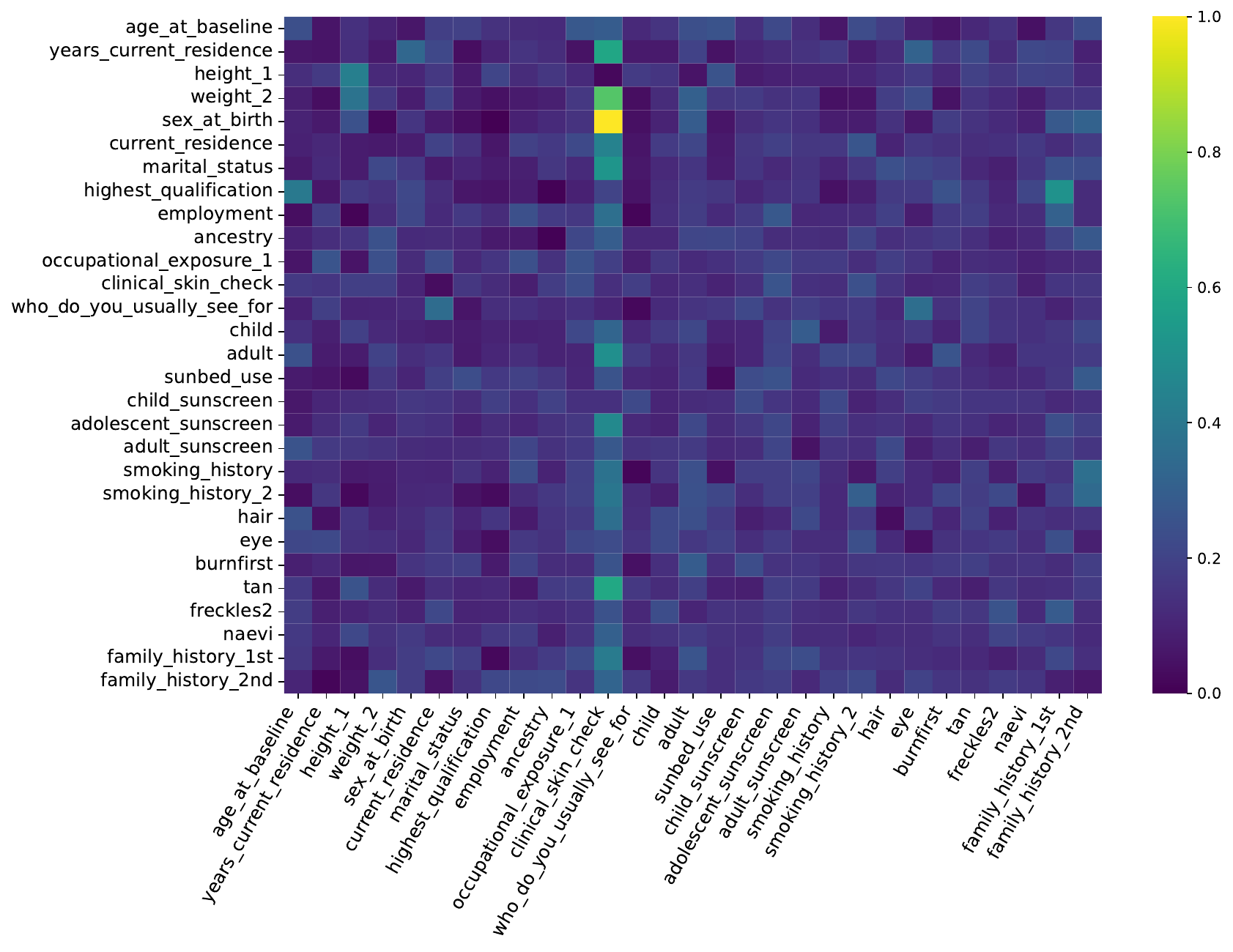} &
\includegraphics[height=6.9cm,trim={0 -2cm 0 0}]{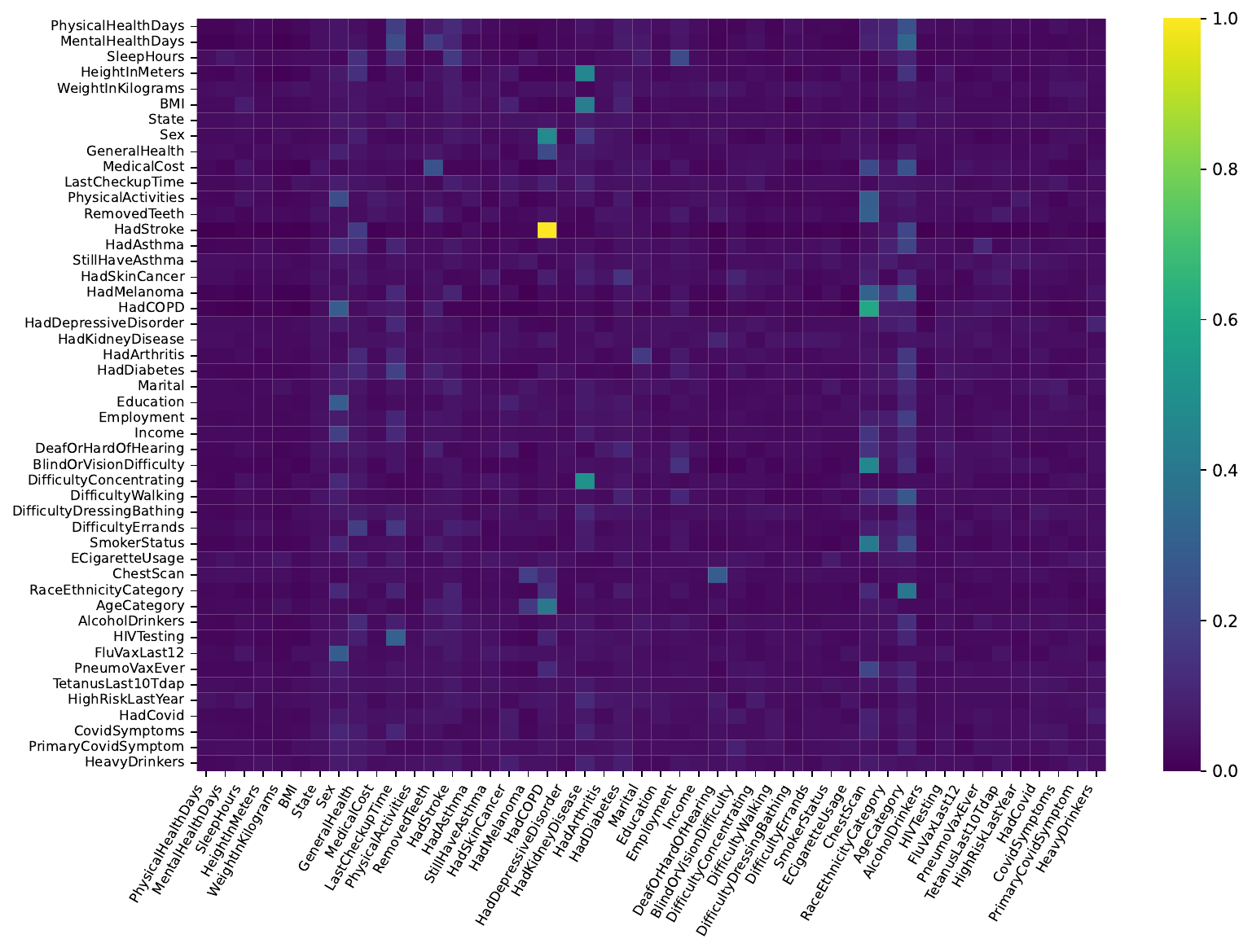}
\end{tabular}
\end{center}
\caption{Attention map visualization generated by the TRACE model's final layer, for the SBMS (left column) and BRFSS~2022 (right column) datasets. The top row demonstrates the attention weights for each feature. For the BRFSS~2022 dataset, 100 random samples are selected from the validation set. The bottom row depicts the relationship between queries (rows) and keys (columns) (Best viewed zoomed-in)} 
\label{fig:attmaps}
\end{figure*}

\subsection{Attention Maps}
Since the TRACE model employs the self-attention mechanism, it offers great insights and interpretability into the decision-making process and results, as visualized by the attention maps in Figure \ref{fig:attmaps}. More specifically, the top row of Figure~\ref{fig:attmaps} represents the attention weight visualizations for the melanoma (left) and heart disease/attack (right) classification tasks. Its aim is to identify the features that predominantly influence the risk estimation. Rows represent participants randomly selected from the validation set and columns the features considered during training. Each cell is calculated by averaging the attention weights of each feature across all the input queries. For instance, the frequency of skin checks a patient has, consistently provides high attention weights across the majority of the participants within the SBMS dataset. Similarly, for the heart attack/disease classification task, the participant's history of CT/CAT scan, COPD (chronic obstructive pulmonary disease), emphysema, chronic bronchitis, stroke and age category are examples of predominant features. In the bottom row of Figure~\ref{fig:attmaps}, each cell represents how much focus to place on each feature when processing a single query feature, revealing underlying relationships between pairs of features. Diagonal elements typically represent self-attention and high values on the diagonal indicate that the model is giving significant importance to individual features.

\subsection{Ablation}

\begin{table}[tbp]
\caption{Ablation on the proposed checkbox embedding feature for the SBMS dataset. CE: checkbox embeddings. }\label{tab:checkbox_embeds_ablation}
\vspace{-0.5cm}
\begin{center}
\resizebox{\columnwidth}{!}{
\begin{tabular}{|c|cccc|}
\hline
CE & Accuracy & F1-Score & Sensitivity & Specificity \\
\hline
\xmark & $84.3\% {\scriptstyle \: \pm 3.1\%}$ & $0.888 {\scriptstyle \: \pm 0.025}$ & $\mathbf{0.954 {\scriptstyle \: \pm 0.025}}$ & $0.632 {\scriptstyle \: \pm 0.070}$\\
\hline
\cmark & $\mathbf{86.0\% {\scriptstyle \: \pm 3.7\%}}$ & $\mathbf{0.900 {\scriptstyle \: \pm 0.027}}$ & $0.941 {\scriptstyle \: \pm 0.018}$ & $\mathbf{0.709 {\scriptstyle \: \pm 0.121}}$\\
\hline
\end{tabular}
}
\end{center}
\vspace{-1em}
\end{table}

Table~\ref{tab:checkbox_embeds_ablation} explores the impact of our proposed checkbox embedding mechanism, on the SBMS dataset, which includes ``checkbox'' features. The first row, represents the scenario where checkbox embeddings are not utilized. In this case, each category within the multiple choice features is considered as an independent binary feature and categorical embeddings are applied normally, without any interactions among the categories included in a single feature. The second row, indicates that incorporating checkbox embeddings, improves the model's performance by capturing interactions within each checkbox feature.

\begin{table}[tbp]
\caption{Comparison of performance metrics on the BRFSS~2022 dataset, with and without considering missing values during training, across different hyperparameter $\alpha$ values. Both cases utilize the same validation/test set.}
\label{tab:missing_vals_ablation}
\vspace{-0.5cm}
\begin{center} 
\resizebox{\columnwidth}{!}{%
\begin{tabular}{|l|c|ccccc|c|}
\hline
Dataset & $\alpha$ & Acc. & F1 & Sens. & Spec. & BA & \# samples \\
\hline
\multirow{5}{*}{\shortstack[l]{BRFSS~2022 \\ w/out missing values}} & 0.5 & 83.4\% & 0.453 & 0.462 & 0.899 & 68.1\% & \multirow{5}{*}{1,745}\\
& 0.6 & 85.7\% & 0.510 & 0.500 & 0.919 & 71.0\% &\\
& 0.7 & 83.7\% & 0.496 & 0.538 & 0.889 & 71.4\% &\\
& 0.8 & 80.2\% & 0.481 & 0.615 & 0.835 & 72.5\% &\\
& 0.9 & 73.6\% & 0.459 & 0.750 & 0.734 & 74.2\% &\\
\hline
\multirow{5}{*}{\shortstack[l]{BRFSS~2022 \\ with missing values}} & 0.5 & \textbf{87.4\%} & 0.488 & 0.404 & \textbf{0.956} & 68.0\% & \multirow{5}{*}{440,111}\\
& 0.6 & \underline{86.5\%} & \underline{0.544} & 0.538 & \underline{0.923} & 73.1\% &\\
& 0.7 & 84.8\% & \textbf{0.555} & 0.635 & 0.886 & 76.1\% &\\
& 0.8 & 77.9\% & 0.528 & \underline{0.827} & 0.771 & \textbf{79.9\%} &\\
& 0.9 & 69.6\% & 0.465 & \textbf{0.885} & 0.663 & \underline{77.4\%} &\\
\hline
\end{tabular}
}
\end{center}
\end{table}

Table~\ref{tab:missing_vals_ablation} compares the performance of our model on the BRFSS~2022 dataset, under two different training scenarios. At first, only samples with complete data are utilized during training (first five rows), resulting in a much smaller dataset with 1,745 samples. On the contrary, the second scenario (last five rows), includes all available samples, by effectively handling missing features, which increases the total number of samples to 440,111. Both scenarios are evaluated with the same validation set. The results suggest that incorporating additional entries, even affected by missing data, enhances the model's ability to generalize, improving performance across all metrics.

\begin{figure}[t!]
\begin{center}
\begin{tabular}{c} 
\begin{tikzpicture}
    \begin{axis}[
        width=\columnwidth, 
        height=6.5cm,  
        xlabel={Missing Values Percentage (\%)},
        ymin=0.55, ymax=0.95,
        xtick={1,10,20,30,40,50,60},
        xticklabels={1,10,20,30,40,50,60},
        ytick={0.6,0.7,0.8,0.9},
        enlargelimits=false,
        grid=major,
        scaled x ticks=false,
        legend pos=south west,
        legend style={font=\tiny}
    ]

\addplot[
    mark=*,
    draw=teal,
    mark size=1.5pt,
    mark options={fill=teal},
]
table [
    x=MissingVals,
    y=BAccuracy,
    col sep=space,
] {
    MissingVals BAccuracy
    1 0.832
    2 0.839
    3 0.771
    4 0.835
    5 0.800
    10 0.720
    20 0.731
    30 0.662
    40 0.633
    50 0.562
};
\addlegendentry{Balanced Accuracy}

\addplot[
    mark=*,
    draw=violet,
    mark size=1.5pt,
    mark options={fill=violet},
]
table [
    x=MissingVals,
    y=f1score,
    col sep=space,
] {
    MissingVals f1score
    1 0.921
    2 0.929
    3 0.898
    4 0.912
    5 0.885
    10 0.894
    20 0.893
    30 0.882
    40 0.840
    50 0.828
};
\addlegendentry{F1-Score}

    \end{axis}
\end{tikzpicture}
\end{tabular}
\end{center}
\caption{Performance of TRACE in terms of F1-Score and Balanced Accuracy in relation to the ratio of missing values on the SBMS dataset.}
\label{fig:missing_vals_simulation}
\end{figure}

Figure~\ref{fig:missing_vals_simulation} illustrates the performance of our proposed method (shown on the \textit{y} axis), across various ratios of randomly simulated missing values (shown on the \textit{x} axis). Balanced Accuracy captures the combined performance of both Specificity and Sensitivity metrics, while F1-Score reflects the overall performance of the trained model. As expected, higher percentages of missing values (up to $50\%$) lead to a decline in performance. F1-Score is not significantly affected when up to $30\%$ of the data is masked out. Balanced Accuracy, however, is more sensitive, exhibiting a descending trend, when the percentage of missing values exceeds $5\%$ of the data.  

Regarding the proposed model, a thorough hyper-parameter optimization has been performed regarding the model size (Table~\ref{tab:model_size_ablation}) and the encoder layers (Table~\ref{tab:enc_layers_ablation}).  Increasing the model size and thus the number of trainable parameters did not lead to significant improvements in performance metrics, leading us to not consider models with sizes greater than 128 from further evaluation. In Table~\ref{tab:enc_layers_ablation}, a similar trend is observed when varying the number of transformer encoder layers; increasing the layers did not result in significant performance improvements. Therefore, the study concluded that focusing on models with 64 and 128 sizes, and utilizing fewer encoder layers, is more efficient without sacrificing accuracy.

\begin{table}[b!]
\caption{Ablation on the TRACE model size for the BRFSS~2022 dataset. All models are trained utilizing the Focal loss, with $\alpha$ hyperparameter equal to $0.8$.}
\label{tab:model_size_ablation}
\vspace{-0.3cm}
\begin{center} 
\resizebox{\columnwidth}{!}{%
\begin{tabular}{|c|ccccc|c|}
\hline
Model Size & Accuracy & F1-Score & Sens. & Spec. & BA & Params \\
\hline
64  & 85.9\% & 0.432 & 0.597 & 0.885 & 74.1\% & 90,497\\
128 & 86.6\% & 0.431 & 0.564 & 0.896 & 73.0\% & 328,449\\
256 & 86.1\% & 0.431 & 0.584 & 0.889 & 73.7\% & 1,246,721\\
512 & 86.1\% & 0.430 & 0.583 & 0.889 & 73.6\% & 4,852,737\\
\hline
\end{tabular}
}
\end{center}
\end{table}

\begin{table}[t!]
\caption{Ablation on the number of transformer encoder layers deployed on the TRACE model for the BRFSS~2022 dataset. The ablation conducted for both 64 and 128 model sizes, while all models are trained utilizing the Focal loss, with $\alpha$ hyperparameter equal to $0.8$.}
\label{tab:enc_layers_ablation}
\vspace{-0.3cm}
\begin{center}
\resizebox{\columnwidth}{!}{%
\begin{tabular}{|c|c|ccccc|c|}
\hline
Model Size & Enc.layers & Acc. & F1 & Sens. & Spec. & BA & Params \\
\hline
\multirow{6}{*}{64} & 1 & 85.9\% & 0.432 & 0.597 & 0.885 & 74.1\% & 90,497\\
& 2 & 86.5\% & 0.430 & 0.566 & 0.894 & 73.0\% & 140,481\\
& 3 & 86.3\% & 0.430 & 0.575 & 0.891 & 73.3\% & 190,465\\
& 4 & 86.5\% & 0.431 & 0.567 & 0.894 & 73.1\% & 240,449\\
& 5 & 86.8\% & 0.430 & 0.553 & 0.899 & 72.6\% & 290,443\\
& 6 & 86.8\% & 0.430 & 0.555 & 0.898 & 72.7\% & 340,417\\
\hline
\multirow{6}{*}{128} & 1 & 86.6\% & 0.431 & 0.564 & 0.896 & 73.0\% & 328,449\\
& 2 & 86.0\% & 0.429 & 0.582 & 0.888 & 73.5\% & 526,721\\
& 3 & 86.3\% & 0.429 & 0.572 & 0.892 & 73.2\% & 724,993\\
& 4 & 86.9\% & 0.428 & 0.545 & 0.901 & 72.3\% & 923,265\\
& 5 & 85.6\% & 0.429 & 0.600 & 0.882 & 74.1\% & 1,121,537\\
& 6 & 84.7\% & 0.428 & 0.636 & 0.868 & 75.2\% & 1,319,809\\
\hline
\end{tabular}
}
\end{center}
\end{table}

\begin{table}[htbp]
\caption{Comparison of imbalance dataset handling techniques, including focal loss, oversampling (O), undersampling (U), and generative approaches on BRFSS-22 dataset.}
\label{tab:results_melanoma_imbalance}
\vspace{-0.5cm}
\begin{center} 
\resizebox{\columnwidth}{!}{%
\begin{tabular}{|l|ccccc|c|}
\hline
Method & Accuracy & F1-Score & Sensitivity & Specificity & BA  \\
\hline
Focal Loss & $86.6\% $ & $\mathbf{0.431} $ & $0.564 $ & $0.896 $ & $73.0\% $\\
\hline
SMOTE (O) & $\mathbf{87.4\%} $ & $0.380 $ & $0.430 $ & $\mathbf{0.918} $ & $67.4\% $\\
\hline
SMOTE (O\&U)  & $84.9\% $ & $0.397 $ & $0.551 $ & $0.879 $ & $71.5\% $\\
\hline
CTGAN & $85.6\% $ & $0.423 $ & $\mathbf{0.588} $ & $0.881 $ & $\mathbf{73.5\%} $\\
\hline
\end{tabular}
}
\end{center}
\end{table}

Since most datasets in our evaluation exhibit class imbalance, a common issue in the medical domain, we adopted several strategies to mitigate the bias toward the benign class. Our primary approach was to use focal loss. However, for the highly imbalanced BRFSS22 dataset, we also explored generative techniques, including SMOTE alone, SMOTE combined with downsampling, and the more robust CTGAN. For SMOTE alone, oversampling the malignant class to 25\% of the benign class (with k=3 nearest neighbors) yielded optimal results, whereas combining SMOTE with a 50\% downsampling of the benign class also proved to be effective. Meanwhile, CTGAN produced high-quality synthetic data that fully balanced the classes (1:1 ratio). We trained for 5000 epochs using a learning rate of $10^{-4}$ for both the generator and discriminator. The network consists of 3 hidden layers of with 256 neurons each and a PAC value of 10 (the number of real samples grouped and fed to the discriminator at once). Overall, as indicated in Table~\ref{tab:results_melanoma_imbalance}, focal loss and CTGAN achieved the best performance in terms of F1-score and Balanced Accuracy, with focal loss being the most efficient during training.

\section{Conclusions}
In this paper, we proposed a novel clinical risk assessment method that utilizes the self-attention mechanism within a Transformer-based architecture. The proposed method, by effectively handling various data modalities and incomplete data, achieves a competitive performance when compared to state-of-the-art methods while requiring significantly less trainable parameters, thus minimizing the computational cost. Additionally, a strong baseline for clinical risk estimation is introduced based on non-negative neural networks (nnMLP), which constrains the network weights to be non-negative, ensuring that the exposure to risk factors only increases the risk of an adverse outcome.

\section*{Acknowledgments}
We gratefully acknowledge the support of NVIDIA Corporation with the donation of the GPUs used for this research.

\bibliographystyle{IEEEtran.bst}
\bibliography{IEEEabrv, biblio.bib}

\appendices

\section{Training Details}\label{sec:app-training}

For training the proposed TRACE model, we employ the LR scheduler ``ReduceLROnPlateau'' which reduces the learning rate when the F1-Score metric (harmonic mean of Precision and Recall of the positive class) stops improving, with $10$ epochs of patience and a decay factor of $0.9$. We choose to monitor F1-Score metric, suitable for both balanced and imbalanced datasets, combining Precision and Recall metrics, for a consistent evaluation protocol.

The deep learning baselines evaluated in this study were deployed and trained via the \textit{PyTorch Tabular framework} \cite{joseph2021pytorch}. For each experiment across all datasets, we utilize the built-in learning rate finder algorithm of Pytorch Tabular, to determine the optimal initial learning rate. Additionally, given the varying nature and dimensionality of the datasets, we perform a hyperparameter search for each method using the \textit{optuna framework}. AdamW was consistently the best performing optimizer across all baselines. As done for TRACE, we employ ``ReduceLROnPlateau'', optimizing the F1-Score metric, with $10$ epochs of patience and a decay factor of $0.9$ for training competing methods also. In Table \ref{tab:pt_hparams} we detail the hyperparameters explored for each baseline within the PyTorch Tabular framework. We preform the hyperparameter search for every dataset and report the best performing setup, to ensure a fair and robust comparison.

\begin{table}[htbp]
\caption{We present the set of hyperparameters and the corresponding search grids used for each baseline within the PyTorch Tabular framework. For any hyperparameters not listed in the table, default values were considered.}
\label{tab:pt_hparams}
\vspace{-0.3cm}
\begin{center} 
\begin{tabular}{lc}
\hline
TabNet & \\
\quad \tt n\_d & $64$\\
\quad \tt n\_a & $64$\\
\quad \tt n\_steps & $[3, 10]$\\
DANET & \\
\quad \tt n\_layers & $\{8, 20, 32\}$\\
\quad \tt abstlay\_dim\_1 & $32$\\
\quad \tt abstlay\_dim\_2 & $64$\\
\quad \tt k & $5$\\
GANDALF & \\
\quad \tt gflu\_stages & $[2,10]$\\
GATE & \\
\quad \tt gflu\_stages & $\{4,6,8\}$\\
\quad \tt num\_trees & $[3,5]$\\
\quad \tt tree\_depth & $\{10,15,20\}$\\
FT-Transformer & \\
\quad \tt input\_embed\_dim & $\{32,64,128\}$\\
\quad \tt num\_heads & $\{4,8\}$\\
\quad \tt num\_attn\_blocks & $\{4,5,6\}$\\
\quad \tt attn\_dropout & $[0.,0.2]$\\
\quad \tt add\_norm\_dropout & $[0.,0.2]$\\
\quad \tt ff\_dropout & $[0.,0.2]$\\
\quad \tt transformer\_activation & $GEGLU$\\
\hline
\end{tabular}
\end{center}
\end{table}

\begin{figure*}[t!]
\begin{center}
\begin{tabular}{cc} 
\begin{tikzpicture}
    \begin{axis}[
        width=8cm, 
        height=6.5cm,  
        xlabel={Number of Parameters},
        ylabel={Balanced Accuracy (\%)},
        ymin=49, ymax=85,
        xtick={10000,300000,500000,1000000,5000000},
        xticklabels={10K,300K,500K,1M,5M},
        ytick={50, 55, 60, 65, 70, 75, 80, 85, 90},
        enlargelimits=true,
        grid=major,
        scaled x ticks=false
    ]

\addplot[
    only marks,
    mark=*,
    draw=black,
    fill=black,
    mark size=1.5pt,
    nodes near coords,
    every node near coord/.append style={font=\tiny,anchor=south west, yshift=-3pt, xshift=-1pt},
    point meta=explicit symbolic, 
    error bars/y explicit,
    error bars/y dir=both,
    error bars/error mark options={draw=black, rotate=90}
]
table [
    meta=Method,
    x=Parameters,
    y=BAccuracy,
    y error=Error,
    col sep=space,
] {
    Method BAccuracy Parameters Error
    TabNet 54.7 698180 8.5
    GATE 73.4 651806 7.0
};

\addplot[
    only marks,
    mark=*,
    draw=black,
    fill=black,
    mark size=1.5pt,
    nodes near coords,
    every node near coord/.append style={font=\tiny,anchor=north east, yshift=0pt, xshift=1pt},
    point meta=explicit symbolic, 
    error bars/y explicit,
    error bars/y dir=both,
    error bars/error mark options={draw=black, rotate=90}
]
table [
    meta=Method,
    x=Parameters,
    y=BAccuracy,
    y error=Error,
    col sep=space,
] {
    Method BAccuracy Parameters Error
    FT-T 61.3 472882 7.7
    DANET 66.9 942074 5.9
    GANDALF 77.1 564000 5.7
};

\addplot[
    only marks,
    mark=*,
    draw=blue,
    fill=blue,
    mark size=1.5pt,
    nodes near coords,
    every node near coord/.append style={font=\tiny,anchor=south west, yshift=-3pt, xshift=-1pt},
    point meta=explicit symbolic, 
    error bars/y explicit,
    error bars/y dir=both,
    error bars/error mark options={draw=blue, rotate=90}
]
table [
    meta=Method,
    x=Parameters,
    y=BAccuracy,
    y error=Error,
    col sep=space,
] {
    Method BAccuracy Parameters Error
    nnMLP 76.5 9665 4.5
};

\addplot[
    only marks,
    mark=*,
    draw=orange,
    fill=orange,
    mark size=1.5pt,
    nodes near coords,
    every node near coord/.append style={font=\tiny,anchor=south west, yshift=-3pt, xshift=-1pt},
    point meta=explicit symbolic, 
    error bars/y explicit,
    error bars/y dir=both,
    error bars/error mark options={draw=orange, rotate=90}
]
table [
    meta=Method,
    x=Parameters,
    y=BAccuracy,
    y error=Error,
    col sep=space,
] {
    Method BAccuracy Parameters Error
    TRACE 82.5 285953 5.3
};

\path [name path=pareto-min]
    (472882, 45)
    -- (472882, 53.6)
    -- (564000, 71.4)
    -- (5000000, 71.4);

\path [name path=pareto-max]
    (472882, 45)
    -- (472882, 69.0)
    -- (564000, 82.8)
    -- (5000000, 82.8);

\path [name path=top]
    (5000000, 100) -- (5000000, 45);

\addplot [fill=purple!60, opacity=0.3]
    fill between [of=pareto-min and top];

\addplot [fill=purple!30, opacity=0.3]
    fill between [of=pareto-max and pareto-min];

\addplot[thick, dashed, purple!60, mark=none] coordinates {
    (472882, 53.6)
    (564000, 71.4)
};

\addplot[thick, dashed, purple!30, mark=none] coordinates {
    (472882, 69.0)
    (564000, 82.8)
};

    \end{axis}
\end{tikzpicture} \quad & \quad
\begin{tikzpicture}
    \begin{axis}[
        width=8cm, 
        height=6.5cm,  
        xlabel={Number of Parameters},
        ymin=62, ymax=76,
        xtick={300000,1000000,4500000,5000000},
        xticklabels={300K,1M,4.5M,5M},
        ytick={60, 65, 70, 75, 80},
        enlargelimits=true,
        grid=major,
        scaled x ticks=false
    ]

\addplot[
    only marks,
    mark=*,
    draw=black,
    fill=black,
    mark size=1.5pt,
    nodes near coords,
    every node near coord/.append style={font=\tiny,anchor=south west, yshift=-2pt, xshift=-2pt},
    point meta=explicit symbolic, 
    error bars/y explicit,
    error bars/y dir=both,
    error bars/error mark options={draw=black, rotate=90}
]
table [
    meta=Method,
    x=Parameters,
    y=BAccuracy,
    y error=Error,
    col sep=space,
] {
    Method BAccuracy Parameters Error
    DANET 72.6 4543161 2.5
    GANDALF 68.0 1044538 6.2
    GATE 72.1 588704 1.9
};

\addplot[
    only marks,
    mark=*,
    draw=black,
    fill=black,
    mark size=1.5pt,
    nodes near coords,
    every node near coord/.append style={font=\tiny,anchor=north east, yshift=1pt, xshift=1pt},
    point meta=explicit symbolic, 
    error bars/y explicit,
    error bars/y dir=both,
    error bars/error mark options={draw=black, rotate=90}
]
table [
    meta=Method,
    x=Parameters,
    y=BAccuracy,
    y error=Error,
    col sep=space,
] {
    Method BAccuracy Parameters Error
    TabNet 71.9 494744 1.2
    FT-T 72.8 4366810 3.5
};

\addplot[
    only marks,
    mark=*,
    draw=blue,
    fill=blue,
    mark size=1.5pt,
    nodes near coords,
    every node near coord/.append style={font=\tiny,anchor=south west, yshift=-3pt, xshift=-1pt},
    point meta=explicit symbolic, 
    error bars/y explicit,
    error bars/y dir=both,
    error bars/error mark options={draw=blue, rotate=90}
]
table [
    meta=Method,
    x=Parameters,
    y=BAccuracy,
    y error=Error,
    col sep=space,
] {
    Method BAccuracy Parameters Error
    nnMLP 75.8 17409 0.3
};

\addplot[
    only marks,
    mark=*,
    draw=orange,
    fill=orange,
    mark size=1.5pt,
    nodes near coords,
    every node near coord/.append style={font=\tiny,anchor=south east, yshift=-3pt, xshift=0pt},
    point meta=explicit symbolic, 
    error bars/y explicit,
    error bars/y dir=both,
    error bars/error mark options={draw=orange, rotate=90}
]
table [
    meta=Method,
    x=Parameters,
    y=BAccuracy,
    y error=Error,
    col sep=space,
] {
    Method BAccuracy Parameters Error
    TRACE 73.1 328449 0.3
};

\path [name path=pareto-min]
    (494744, 55)
    -- (494744, 70.7)
    -- (5000000, 70.7);

\path [name path=pareto-max]
    (494744, 55)
    -- (494744, 73.1)
    -- (588704, 74.0)
    -- (1044538, 74.2)
    -- (4366810, 76.3)
    -- (5000000, 76.3);

\path [name path=top]
    (5000000, 100) -- (5000000, 45);

\addplot [fill=purple!60, opacity=0.3]
    fill between [of=pareto-min and top];

\addplot [fill=purple!30, opacity=0.3]
    fill between [of=pareto-max and pareto-min];


\addplot[thick, dashed, purple!30, mark=none] coordinates {
    (494744, 73.1)
    (588704, 74.0)
    (1044538, 74.2)
    (4366810, 76.3)
};

    \end{axis}
\end{tikzpicture}
\end{tabular}
\end{center}
\caption{Balanced Accuracy vs. number of trainable parameters on SBMS (left) and BRFSS-2022 (right) datasets. Results are shown for the proposed \textcolor{blue}{nnMLP} and \textcolor{orange}{TRACE} models against TabNet, DANET, FT-Transformer, GANDALF and GATE.} 
\label{fig:acc_vs_params_itobos_brfss}
\end{figure*}

\section{Dataset Properties}\label{sec:app1}

To convert the original multiclass labels into binary categories, we grouped the lesions as follows:

PAD-UFES-20: Lesions were labeled as malignant if they were Basal Cell Carcinoma (BCC), Melanoma (MEL), or Squamous Cell Carcinoma (SCC), lesions classified as Actinic Keratosis (ACK), Nevus (NEV), or Seborrheic Keratosis (SEK) were considered benign.

HIBA: Malignancy was assigned to BCC, MEL, and SCC, while benign lesions included Actinic Keratosis (ACK), Dermatofibroma (DF), Lichenoid Keratosis (LK), Seborrheic Keratosis (SEK), Nevus (NEV), Vascular Lesion (VASC), and Solar Lentigo (SL).

\added{HAM10000: Lesions identified as BCC or MEL were deemed malignant, and those labeled as Actinic Keratosis (ACK), Nevus (NEV), Vascular Lesion (VASC), 
Dermatofibroma (DF), or Benign Keratosis-like Lesions (BKL) were categorized as benign.}

Regarding the datasets, we encountered varying levels of missing values. HAM10000 contain relatively low percentages of missing values at 0.5\%. On the contrary, PAD-UFES-20, HIBA and BRFSS-2022 have a higher ratio of missing data, with 17.55\%, 11.34\% and 12.7\% respectively. These percentages were calculated across all features for each dataset. Notably, SBMS dataset, although it includes fewer samples, does not contain any missing values.

\begin{figure*}[t!]
\begin{center}
\begin{tabular}{cc} 
\begin{tikzpicture}
    \begin{axis}[
        width=8cm, 
        height=6.5cm,  
        ylabel={Balanced Accuracy (\%)},
        xmode=log,
        log basis x=10,
        ymin=79, ymax=94,
        xtick={10000,300000,1000000,3000000},
        xticklabels={10K,300K,1M,3M},
        ytick={80, 85, 90, 95},
        enlargelimits=true,
        grid=major,
        scaled x ticks=false
    ]

\addplot[
    only marks,
    mark=*,
    draw=black,
    fill=black,
    mark size=1.5pt,
    nodes near coords,
    every node near coord/.append style={font=\tiny,anchor=south west, yshift=-3pt, xshift=-1pt},
    point meta=explicit symbolic, 
    error bars/y explicit,
    error bars/y dir=both,
    error bars/error mark options={draw=black, rotate=90}
]
table [
    meta=Method,
    x=Parameters,
    y=BAccuracy,
    y error=Error,
    col sep=space,
] {
    Method BAccuracy Parameters Error
    FT-T 83.5 2771698 0.7
    DANET 86.0 1684853 3.0
    GATE 85.6 505692 3.0
};

\addplot[
    only marks,
    mark=*,
    draw=black,
    fill=black,
    mark size=1.5pt,
    nodes near coords,
    every node near coord/.append style={font=\tiny,anchor=south east, yshift=-3pt, xshift=0pt},
    point meta=explicit symbolic, 
    error bars/y explicit,
    error bars/y dir=both,
    error bars/error mark options={draw=black, rotate=90}
]
table [
    meta=Method,
    x=Parameters,
    y=BAccuracy,
    y error=Error,
    col sep=space,
] {
    Method BAccuracy Parameters Error
    TabNet 83.2 392668 0.7
    GANDALF 80.9 169554 2.3
};

\addplot[
    only marks,
    mark=*,
    draw=blue,
    fill=blue,
    mark size=1.5pt,
    nodes near coords,
    every node near coord/.append style={font=\tiny,anchor=south west, yshift=-3pt, xshift=-1pt},
    point meta=explicit symbolic, 
    error bars/y explicit,
    error bars/y dir=both,
    error bars/error mark options={draw=blue, rotate=90}
]
table [
    meta=Method,
    x=Parameters,
    y=BAccuracy,
    y error=Error,
    col sep=space,
] {
    Method BAccuracy Parameters Error
    nnMLP 90.8 8129 1.5
};

\addplot[
    only marks,
    mark=*,
    draw=orange,
    fill=orange,
    mark size=1.5pt,
    nodes near coords,
    every node near coord/.append style={font=\tiny,anchor=south west, yshift=-3pt, xshift=-1pt},
    point meta=explicit symbolic, 
    error bars/y explicit,
    error bars/y dir=both,
    error bars/error mark options={draw=orange, rotate=90}
]
table [
    meta=Method,
    x=Parameters,
    y=BAccuracy,
    y error=Error,
    col sep=space,
] {
    Method BAccuracy Parameters Error
    TRACE 92.4 674945 2.8
};

\path [name path=pareto-min]
    (169554, 75)
    -- (169554, 78.6)
    -- (392668, 82.5)
    -- (505692, 82.6)
    -- (1684853, 83.0)
    -- (5000000, 83.0);

\path [name path=pareto-max]
    (169554, 75)
    -- (169554, 83.2)
    -- (505692, 88.6)
    -- (1684853, 89.0)
    -- (5000000, 89.0);

\path [name path=top]
    (5000000, 97) -- (5000000, 75);

\addplot [fill=purple!60, opacity=0.3]
    fill between [of=pareto-min and top];

\addplot [fill=purple!30, opacity=0.3]
    fill between [of=pareto-max and pareto-min];

\addplot[thick, dashed, purple!60, mark=none] coordinates {
    (169554, 78.6)
    (392668, 82.5)
    (505692, 82.6)
    (1684853, 83.0)
};

\addplot[thick, dashed, purple!30, mark=none] coordinates {
    (169554, 83.2)
    (505692, 88.6)
    (1684853, 89.0)
};

    \end{axis}
\end{tikzpicture} \quad & \quad
\begin{tikzpicture}
    \begin{axis}[
        width=8cm, 
        height=6.5cm,  
        ylabel={Accuracy (\%)},
        xmode=log,
        log basis x=10,
        ymin=62, ymax=84,
        xtick={10000,300000,1000000,3000000},
        xticklabels={10K,300K,1M,3M},
        ytick={55, 60, 65, 70, 75, 80, 85},
        enlargelimits=true,
        grid=major,
        scaled x ticks=false
    ]

\addplot[
    only marks,
    mark=*,
    draw=black,
    fill=black,
    mark size=1.5pt,
    nodes near coords,
    every node near coord/.append style={font=\tiny,anchor=north east, yshift=3pt, xshift=0pt},
    point meta=explicit symbolic, 
    error bars/y explicit,
    error bars/y dir=both,
    error bars/error mark options={draw=black, rotate=90}
]
table [
    meta=Method,
    x=Parameters,
    y=Accuracy,
    y error=Error,
    col sep=space,
] {
    Method Accuracy Parameters Error
    TabNet 62.3 392917 1.2 
    DANET 75.9 1685106 0.6 
    GANDALF 78.7 169791 3.0 
    GATE 75.6 505817 3.9 
};

\addplot[
    only marks,
    mark=*,
    draw=black,
    fill=black,
    mark size=1.5pt,
    nodes near coords,
    every node near coord/.append style={font=\tiny,anchor=south west, yshift=-3pt, xshift=-1pt},
    point meta=explicit symbolic, 
    error bars/y explicit,
    error bars/y dir=both,
    error bars/error mark options={draw=black, rotate=90}
]
table [
    meta=Method,
    x=Parameters,
    y=Accuracy,
    y error=Error,
    col sep=space,
] {
    Method Accuracy Parameters Error
    FT-T 75.1 2772086 1.4
};

\addplot[
    only marks,
    mark=*,
    draw=blue,
    fill=blue,
    mark size=1.5pt,
    nodes near coords,
    every node near coord/.append style={font=\tiny,anchor=south west, yshift=-3pt, xshift=-1pt},
    point meta=explicit symbolic, 
    error bars/y explicit,
    error bars/y dir=both,
    error bars/error mark options={draw=blue, rotate=90}
]
table [
    meta=Method,
    x=Parameters,
    y=Accuracy,
    y error=Error,
    col sep=space,
] {
    Method Accuracy Parameters Error
    nnMLP 79.4 8294 2.2 
};

\addplot[
    only marks,
    mark=*,
    draw=orange,
    fill=orange,
    mark size=1.5pt,
    nodes near coords,
    every node near coord/.append style={font=\tiny,anchor=south west, yshift=-3pt, xshift=-1pt},
    point meta=explicit symbolic, 
    error bars/y explicit,
    error bars/y dir=both,
    error bars/error mark options={draw=orange, rotate=90}
]
table [
    meta=Method,
    x=Parameters,
    y=Accuracy,
    y error=Error,
    col sep=space,
] {
    Method Accuracy Parameters Error
    TRACE 83.2 675590 2.1
};

\path [name path=pareto-min]
    (169791, 55)
    -- (169791, 75.7)
    -- (5000000, 75.7);

\path [name path=pareto-max]
    (169791, 55)
    -- (169791, 81.7)
    -- (5000000, 81.7);

\path [name path=top]
    (5000000, 90) -- (5000000, 55);

\addplot [fill=purple!60, opacity=0.3]
    fill between [of=pareto-min and top];

\addplot [fill=purple!30, opacity=0.3]
    fill between [of=pareto-max and pareto-min];



    \end{axis}
\end{tikzpicture}\\
\begin{tikzpicture}
    \begin{axis}[
        width=8cm, 
        height=6.5cm,  
        ylabel={Balanced Accuracy (\%)},
        ymin=80, ymax=95,
        xtick={10000,500000,1000000,1500000},
        xticklabels={10K,500K,1M,1.5M},
        ytick={80, 85, 90, 95},
        enlargelimits=true,
        grid=major,
        scaled x ticks=false
    ]

\addplot[
    only marks,
    mark=*,
    draw=black,
    fill=black,
    mark size=1.5pt,
    nodes near coords,
    every node near coord/.append style={font=\tiny,anchor=south west, yshift=-1pt, xshift=-2pt},
    point meta=explicit symbolic, 
    error bars/y explicit,
    error bars/y dir=both,
    error bars/error mark options={draw=black, rotate=90}
]
table [
    meta=Method,
    x=Parameters,
    y=BAccuracy,
    y error=Error,
    col sep=space,
] {
    Method BAccuracy Parameters Error
    TabNet 84.5 789940 1.9
    DANET 84.8 1221654 1.1
    GATE 86.7 36722 0.9
};

\addplot[
    only marks,
    mark=*,
    draw=black,
    fill=black,
    mark size=1.5pt,
    nodes near coords,
    every node near coord/.append style={font=\tiny,anchor=north, yshift=-6pt, xshift=5pt},
    point meta=explicit symbolic, 
    error bars/y explicit,
    error bars/y dir=both,
    error bars/error mark options={draw=black, rotate=90}
]
table [
    meta=Method,
    x=Parameters,
    y=BAccuracy,
    y error=Error,
    col sep=space,
] {
    Method BAccuracy Parameters Error
    GANDALF 86.3 7282 1.0
};

\addplot[
    only marks,
    mark=*,
    draw=black,
    fill=black,
    mark size=1.5pt,
    nodes near coords,
    every node near coord/.append style={font=\tiny,anchor=north east, yshift=1pt, xshift=1pt},
    point meta=explicit symbolic, 
    error bars/y explicit,
    error bars/y dir=both,
    error bars/error mark options={draw=black, rotate=90}
]
table [
    meta=Method,
    x=Parameters,
    y=BAccuracy,
    y error=Error,
    col sep=space,
] {
    Method BAccuracy Parameters Error
    FT-T 84.9 1843882 1.4
};

\addplot[
    only marks,
    mark=*,
    draw=blue,
    fill=blue,
    mark size=1.5pt,
    nodes near coords,
    every node near coord/.append style={font=\tiny,anchor=south west, yshift=-3pt, xshift=-1pt},
    point meta=explicit symbolic, 
    error bars/y explicit,
    error bars/y dir=both,
    error bars/error mark options={draw=blue, rotate=90}
]
table [
    meta=Method,
    x=Parameters,
    y=BAccuracy,
    y error=Error,
    col sep=space,
] {
    Method BAccuracy Parameters Error
    nnMLP 91.3 4417 1.4
};

\addplot[
    only marks,
    mark=*,
    draw=orange,
    fill=orange,
    mark size=1.5pt,
    nodes near coords,
    every node near coord/.append style={font=\tiny,anchor=south, yshift=-3pt, xshift=10pt},
    point meta=explicit symbolic, 
    error bars/y explicit,
    error bars/y dir=both,
    error bars/error mark options={draw=orange, rotate=90}
]
table [
    meta=Method,
    x=Parameters,
    y=BAccuracy,
    y error=Error,
    col sep=space,
] {
    Method BAccuracy Parameters Error
    TRACE 91.6 633089 0.8
};

\path [name path=pareto-min]
    (7282, 70)
    -- (7282, 85.3)
    -- (36722, 85.8)
    -- (3000000, 85.8);

\path [name path=pareto-max]
    (7282, 70)
    -- (7282, 87.3)
    -- (36722, 87.6)
    -- (3000000, 87.6);

\path [name path=top]
    (3000000, 100) -- (3000000, 50);

\addplot [fill=purple!60, opacity=0.3]
    fill between [of=pareto-min and top];

\addplot [fill=purple!30, opacity=0.3]
    fill between [of=pareto-max and pareto-min];



    \end{axis}
\end{tikzpicture} \quad & \quad
\begin{tikzpicture}
    \begin{axis}[
        width=8cm, 
        height=6.5cm,  
        ylabel={Accuracy (\%)},
        ymin=51, ymax=70,
        xtick={10000,500000,1000000,1500000},
        xticklabels={10K,500K,1M,1.5M},
        ytick={45, 50, 55, 60, 65, 70},
        enlargelimits=true,
        grid=major,
        scaled x ticks=false
    ]

\addplot[
    only marks,
    mark=*,
    draw=black,
    fill=black,
    mark size=1.5pt,
    nodes near coords,
    every node near coord/.append style={font=\tiny,anchor=south west, yshift=-3pt, xshift=-1pt},
    point meta=explicit symbolic, 
    error bars/y explicit,
    error bars/y dir=both,
    error bars/error mark options={draw=black, rotate=90}
]
table [
    meta=Method,
    x=Parameters,
    y=Accuracy,
    y error=Error,
    col sep=space,
] {
    Method Accuracy Parameters Error
    TabNet 54.8 790452 3.9
    FT-T 65.6 1844914 1.7
    DANET 64.8 1222174 2.5
    GANDALF 67.5 7490 1.8
};

\addplot[
    only marks,
    mark=*,
    draw=black,
    fill=black,
    mark size=1.5pt,
    nodes near coords,
    every node near coord/.append style={font=\tiny,anchor=north west, yshift=3pt, xshift=-1pt},
    point meta=explicit symbolic, 
    error bars/y explicit,
    error bars/y dir=both,
    error bars/error mark options={draw=black, rotate=90}
]
table [
    meta=Method,
    x=Parameters,
    y=Accuracy,
    y error=Error,
    col sep=space,
] {
    Method Accuracy Parameters Error
    GATE 64.4 36794 2.0
};

\addplot[
    only marks,
    mark=*,
    draw=blue,
    fill=blue,
    mark size=1.5pt,
    nodes near coords,
    every node near coord/.append style={font=\tiny,anchor=north east, yshift=2pt, xshift=1pt},
    point meta=explicit symbolic, 
    error bars/y explicit,
    error bars/y dir=both,
    error bars/error mark options={draw=blue, rotate=90}
]
table [
    meta=Method,
    x=Parameters,
    y=Accuracy,
    y error=Error,
    col sep=space,
] {
    Method Accuracy Parameters Error
    nnMLP 64.5 4714 2.8
};

\addplot[
    only marks,
    mark=*,
    draw=orange,
    fill=orange,
    mark size=1.5pt,
    nodes near coords,
    every node near coord/.append style={font=\tiny,anchor=south west, yshift=-3pt, xshift=-1pt},
    point meta=explicit symbolic, 
    error bars/y explicit,
    error bars/y dir=both,
    error bars/error mark options={draw=orange, rotate=90}
]
table [
    meta=Method,
    x=Parameters,
    y=Accuracy,
    y error=Error,
    col sep=space,
] {
    Method Accuracy Parameters Error
    TRACE 67.9 634250 2.2
};

\path [name path=pareto-min]
    (7490, 45)
    -- (7490, 65.7)
    -- (3000000, 65.7);

\path [name path=pareto-max]
    (7490, 45)
    -- (7490, 69.3)
    -- (3000000, 69.3);

\path [name path=top]
    (3000000, 100) -- (3000000, 45);

\addplot [fill=purple!60, opacity=0.3]
    fill between [of=pareto-min and top];

\addplot [fill=purple!30, opacity=0.3]
    fill between [of=pareto-max and pareto-min];

    \end{axis}
\end{tikzpicture}\\

\begin{tikzpicture}
    \begin{axis}[
        width=8cm, 
        height=6.5cm,  
        xlabel={Number of Parameters},
        ylabel={Balanced Accuracy (\%)},
        ymin=61, ymax=83,
        xtick={10000,200000,500000,1000000,1500000},
        xticklabels={10K,200K,500K,1M,1.5M},
        ytick={60, 65, 70, 75, 80, 85},
        enlargelimits=true,
        grid=major,
        scaled x ticks=false
    ]

\addplot[
    only marks,
    mark=*,
    draw=black,
    fill=black,
    mark size=1.5pt,
    nodes near coords,
    every node near coord/.append style={font=\tiny,anchor=south west, yshift=-3pt, xshift=-1pt},
    point meta=explicit symbolic, 
    error bars/y explicit,
    error bars/y dir=both,
    error bars/error mark options={draw=black, rotate=90}
]
table [
    meta=Method,
    x=Parameters,
    y=BAccuracy,
    y error=Error,
    col sep=space,
] {
    Method BAccuracy Parameters Error
    TabNet 74.0 647843 7.0
    FT-T 75.6 462010 8.3
    GANDALF 65.6 3011 5.9
};

\addplot[
    only marks,
    mark=*,
    draw=black,
    fill=black,
    mark size=1.5pt,
    nodes near coords,
    every node near coord/.append style={font=\tiny,anchor=north east, yshift=3pt, xshift=0pt},
    point meta=explicit symbolic, 
    error bars/y explicit,
    error bars/y dir=both,
    error bars/error mark options={draw=black, rotate=90}
]
table [
    meta=Method,
    x=Parameters,
    y=BAccuracy,
    y error=Error,
    col sep=space,
] {
    Method BAccuracy Parameters Error
    GATE 82.5 1033629 1.3
    DANET 80.8 431680 1.0
};

\addplot[
    only marks,
    mark=*,
    draw=blue,
    fill=blue,
    mark size=1.5pt,
    nodes near coords,
    every node near coord/.append style={font=\tiny,anchor=south west, yshift=-3pt, xshift=-1pt},
    point meta=explicit symbolic, 
    error bars/y explicit,
    error bars/y dir=both,
    error bars/error mark options={draw=blue, rotate=90}
]
table [
    meta=Method,
    x=Parameters,
    y=BAccuracy,
    y error=Error,
    col sep=space,
] {
    Method BAccuracy Parameters Error
    nnMLP 78.7 3713 1.8
};

\addplot[
    only marks,
    mark=*,
    draw=orange,
    fill=orange,
    mark size=1.5pt,
    nodes near coords,
    every node near coord/.append style={font=\tiny,anchor=south west, yshift=-3pt, xshift=-1pt},
    point meta=explicit symbolic, 
    error bars/y explicit,
    error bars/y dir=both,
    error bars/error mark options={draw=orange, rotate=90}
]
table [
    meta=Method,
    x=Parameters,
    y=BAccuracy,
    y error=Error,
    col sep=space,
] {
    Method BAccuracy Parameters Error
    TRACE 80.7 160065 1.3
};

\path [name path=pareto-min]
    (3011, 55)
    -- (3011, 59.7)
    -- (431680, 79.8)
    -- (1033629, 81.2)
    -- (2000000, 81.2);

\path [name path=pareto-max]
    (3011, 55)
    -- (3011, 71.5)
    -- (431680, 81.8)
    -- (462010, 83.9)
    -- (2000000, 83.9);

\path [name path=top]
    (2000000, 100) -- (2000000, 55);

\addplot [fill=purple!60, opacity=0.3]
    fill between [of=pareto-min and top];

\addplot [fill=purple!30, opacity=0.3]
    fill between [of=pareto-max and pareto-min];

\addplot[thick, dashed, purple!60, mark=none] coordinates {
    (3011, 59.7)
    (431680, 79.8)
    (1033629, 81.2)
};

\addplot[thick, dashed, purple!30, mark=none] coordinates {
    (3011, 71.5)
    (431680, 81.8)
    (462010, 83.9)
};

    \end{axis}
\end{tikzpicture} \quad & \quad

\begin{tikzpicture}
    \begin{axis}[
        width=8cm, 
        height=6.5cm,  
        xlabel={Number of Parameters},
        ylabel={Accuracy (\%)},
        ymin=69, ymax=75,
        xtick={10000,200000,500000,1000000,1500000},
        xticklabels={10K,200K,500K,1M,1.5M},
        ytick={65, 70, 73, 75},
        enlargelimits=true,
        grid=major,
        scaled x ticks=false
    ]

\addplot[
    only marks,
    mark=*,
    draw=black,
    fill=black,
    mark size=1.5pt,
    nodes near coords,
    every node near coord/.append style={font=\tiny,anchor=south west, yshift=-4pt, xshift=-1pt},
    point meta=explicit symbolic, 
    error bars/y explicit,
    error bars/y dir=both,
    error bars/error mark options={draw=black, rotate=90}
]
table [
    meta=Method,
    x=Parameters,
    y=Accuracy,
    y error=Error,
    col sep=space,
] {
    Method Accuracy Parameters Error
    TabNet 71.4 648163 0.2 
    FT-T 70.3 462335 0.2
    DANET 71.9 432005 0.5
    GANDALF 71.4 3096 0.8
};

\addplot[
    only marks,
    mark=*,
    draw=black,
    fill=black,
    mark size=1.5pt,
    nodes near coords,
    every node near coord/.append style={font=\tiny,anchor=south east, yshift=-3pt, xshift=0pt},
    point meta=explicit symbolic, 
    error bars/y explicit,
    error bars/y dir=both,
    error bars/error mark options={draw=black, rotate=90}
]
table [
    meta=Method,
    x=Parameters,
    y=Accuracy,
    y error=Error,
    col sep=space,
] {
    Method Accuracy Parameters Error
    GATE 70.6 1033794 0.6
};

\addplot[
    only marks,
    mark=*,
    draw=blue,
    fill=blue,
    mark size=1.5pt,
    nodes near coords,
    every node near coord/.append style={font=\tiny,anchor=south west, yshift=-3pt, xshift=-1pt},
    point meta=explicit symbolic, 
    error bars/y explicit,
    error bars/y dir=both,
    error bars/error mark options={draw=blue, rotate=90}
]
table [
    meta=Method,
    x=Parameters,
    y=Accuracy,
    y error=Error,
    col sep=space,
] {
    Method Accuracy Parameters Error
    nnMLP 70.0 3911 0.8 
};

\addplot[
    only marks,
    mark=*,
    draw=orange,
    fill=orange,
    mark size=1.5pt,
    nodes near coords,
    every node near coord/.append style={font=\tiny,anchor=south west, yshift=-3pt, xshift=-1pt},
    point meta=explicit symbolic, 
    error bars/y explicit,
    error bars/y dir=both,
    error bars/error mark options={draw=orange, rotate=90}
]
table [
    meta=Method,
    x=Parameters,
    y=Accuracy,
    y error=Error,
    col sep=space,
] {
    Method Accuracy Parameters Error
    TRACE 73.5 160455 1.8
};

\path [name path=pareto-min]
    (3096, 60)
    -- (3096, 70.6)
    -- (432005, 71.4)
    -- (1500000, 71.4);

\path [name path=pareto-max]
    (3096, 60)
    -- (3096, 72.2)
    -- (432005, 72.4)
    -- (1500000, 72.4);

\path [name path=top]
    (1500000, 100) -- (1500000, 60);

\addplot [fill=purple!60, opacity=0.3]
    fill between [of=pareto-min and top];

\addplot [fill=purple!30, opacity=0.3]
    fill between [of=pareto-max and pareto-min];

\addplot[thick, dashed, purple!60, mark=none] coordinates {
    (3096, 70.6)
    (432005, 71.4)
};

\addplot[thick, dashed, purple!30, mark=none] coordinates {
    (3096, 72.2)
    (432005, 72.4)
};

    \end{axis}
\end{tikzpicture}
\end{tabular}
\end{center}
\caption{Balanced Accuracy vs. number of trainable parameters for the binary classification task (left column) and Accuracy vs. number of trainable parameters for the multiclass classification task (right column). Results are shown for the proposed \textcolor{blue}{nnMLP} and \textcolor{orange}{TRACE} models against TabNet, DANET, FT-Transformer, GANDALF and GATE, on PAD-UFES-20 (top row), HIBA (middle row) and HAM10000 (bottom row) datasets.} 
\label{fig:acc_vs_params}
\end{figure*}

\section{Focal Loss}\label{sec:app2}

Mathematically, the Focal loss is defined as: 
\begin{equation}\label{eq:focal}
    FocalLoss(p_t) = -\alpha_t \cdot (1-p_t)^\gamma \cdot log(p_t),
\end{equation}
where $p_t$ represents the probability corresponding to the true class (disease existence).

It incorporates a modulating factor $(1-p_t)^{\gamma}$ on top of the cross entropy loss criterion. This factor forces the model to focus on hard examples, during training. For $\gamma=0$, the Focal loss is equivalent to Cross Entropy, but for $\gamma>0$ it raises the confidence and subsequently down-weights the contribution of easy examples. In practice, the factor $\alpha$ is used as a weighting factor to balance the contribution from both classes, where $0<\alpha<1$. Setting $\alpha$ near $0$ increases the influence of the negative class, while setting it near $1$ the positive class will contribute more to the final result, despite being less represented. The latter scenario aligns with the objective of improving classification of rarely observed instances.

\section{Complexity Visualization}\label{sec:app-balance-acc}

Figure \ref{fig:acc_vs_params_itobos_brfss} demonstrates the performance of the proposed nnMLP and TRACE models in comparison to baseline methods, with respect to the total number of trainable parameters, for the SBMS and BRFSS-2022 datasets. Each baseline is represented by the average value of the corresponding metric, along with error bars, indicating its standard deviation. Additionally we overlay two Pareto frontiers, defined by the minimum and maximum values, taking into account the standard deviation of the reported metric accross the existing baseline methods, in order to highlight the current optimal trade-off between performance and model complexity.

Both nnMLP and TRACE models, showcase an optimal trade-off between model complexity and performance, since they are consistently located at the top-left of the Pareto frontiers defined by the existing baselines. More specifically, in the SBMS dataset we observe that nnMLP achieves a competitive 76.5\% BA, compared with GANDALF (77.1\%), introducing less than 10,000 trainable parameters, while the best performing GANDALF model has 564,000 trainable parameters. Meanwhile, TRACE achieves a state-of-the-art 82.5\% BA with less than 300,000 parameters. Regarding BRFSS-2022 dataset, the proposed nnMLP baseline model achieves a state-of-the-art 75.8\% BA, with a significant drop of model complexity to just 17,409 parameters.

Similarly, Figure \ref{fig:acc_vs_params} depicts the performance of our proposed models, across PAD-UFES-20, HIBA and HAM10000 datasets in both binary and multiclass setups. The left column depicts the results obtained for the binary classification task under the \textit{malignant vs. benign} setup, thus the Balanced Accuracy metric is reported. The right column presents results for the multiclass classification task, evaluated using the Accuracy metric. Each row corresponds to a different dataset; PAD-UFES-20 (top), HIBA (middle) and HAM10000 (bootom).

We observe that both nnMLP and TRACE are, in most cases, located at the left, top or ideally the top-left of the Pareto frontier, primarily defined by GANDALF, DANET and GATE baselines. Among these, GANDALF is one of the most efficient and consistent baselines in terms of model complexity but with relatively poor performance across the evaluated datasets. The nnMLP baseline model matches or reduces the number of trainable parameters introduced by GANDALF, while achieving better performance in every binary setup, as well as in the multiclass classification setup of PAD-UFES-20 dataset. TRACE, on the other hand, introduces more trainable parameters, but consistently outperforms all baselines as illustrated in Figure \ref{fig:acc_vs_params}. Even in cases that TRACE performs slightly worse, such as the binary setup of HAM10000 dataset compared to GATE, it still demonstrates significantly better trade-off between BA and model complexity.

\section{Limitations}\label{sec:app-limitations}
While our proposed models demonstrate robust and consistent performance across various clinical tasks and multiple evaluated datasets, several limitations remain for discussion, that could be addressed in future work. Developing a large, generalizable pretrained model for clinical data is challenging, as the architectural design of TRACE aligns with the nature of the input features. Additionally, datasets that contain under-represented demographic or socio-economic groups may insert biases in the trained models, potentially impacting generalizability. Finally, TRACE does not explicitly handles temporal information. Although we offer workarounds (e.g., encoding time-steps as separate entries or checkbox style features), these do not capture possible trends over a specific measurement. Modeling temporal data can be achieved by a decoder-only architecture, deploying the masked attention mechanism, which is an important direction for future work.

\newpage

\end{document}